%% file: icml_main.tex
\documentclass{article}

\usepackage{microtype}
\usepackage{graphicx}
\usepackage{subcaption}
\usepackage{booktabs} 

\usepackage{hyperref}

\usepackage[preprint]{icml2026}

\usepackage{amsmath}
\usepackage{amssymb}
\usepackage{mathtools}
\usepackage{amsthm}

\usepackage{url,enumitem}
\usepackage{xspace}
\usepackage{comment}
\usepackage[normalem]{ulem}
\usepackage{mathtools}
\usepackage{svg}
\usepackage{algpseudocode}
\usepackage{dsfont}

\usepackage{multirow}
\usepackage{siunitx}
\usepackage{float}
\usepackage[overload]{ragged2e}
\usepackage{pgfplots}
\usepackage{arydshln}
\usepackage{makecell}
\usepackage{pifont}
\usepackage{pgfplotstable}
\usetikzlibrary{intersections}
\usepgfplotslibrary{fillbetween}

\usepackage{caption}
\usepackage{bm}
\usepackage{wrapfig}
\usepackage{xcolor}

\usepackage[capitalize,noabbrev]{cleveref}

\theoremstyle{plain}

\theoremstyle{definition}

\theoremstyle{remark}

\usepackage[disable,textsize=tiny]{todonotes}

\definecolor{lightblue}{rgb}{0.1,0.3,1}
\definecolor{lightgreen}{rgb}{0.1,0.8,0.1}
\definecolor{defaultcolor}{gray}{0.9}
\newcommand{\std}[1]{{\tiny\textcolor{green!60!black}{$\pm$#1}}}

\icmltitlerunning{When Rubrics Fail: Error Enumeration as Reward in Reference-Free RL Post-Training for Virtual Try-On}

\begin{document}

\title{When Rubrics Fail: Error Enumeration as Reward in Reference-Free RL Post-Training for Virtual Try-On}


\author{
Wisdom Ikezogwo\textsuperscript{1,2}\thanks{Work done during internship at Amazon.} 
\quad
Mehmet Saygin Seyfioglu\textsuperscript{2}\thanks{Equal contribution.}
\quad
Karim Bouyarmane\textsuperscript{2} \vspace{8pt}\\
\textsuperscript{1} University of Washington \quad
\textsuperscript{2}Amazon \quad
\vspace{8pt}\\
{\tt\small wisdomik@uw.edu} \\
{\tt\small \{mseyfiog,bouykari\}@amazon.com}\\
}


\newsavebox{\teaserbox}
\savebox{\teaserbox}{%
  \centering
  \includegraphics[width=1.8\linewidth, height=0.305\textheight]{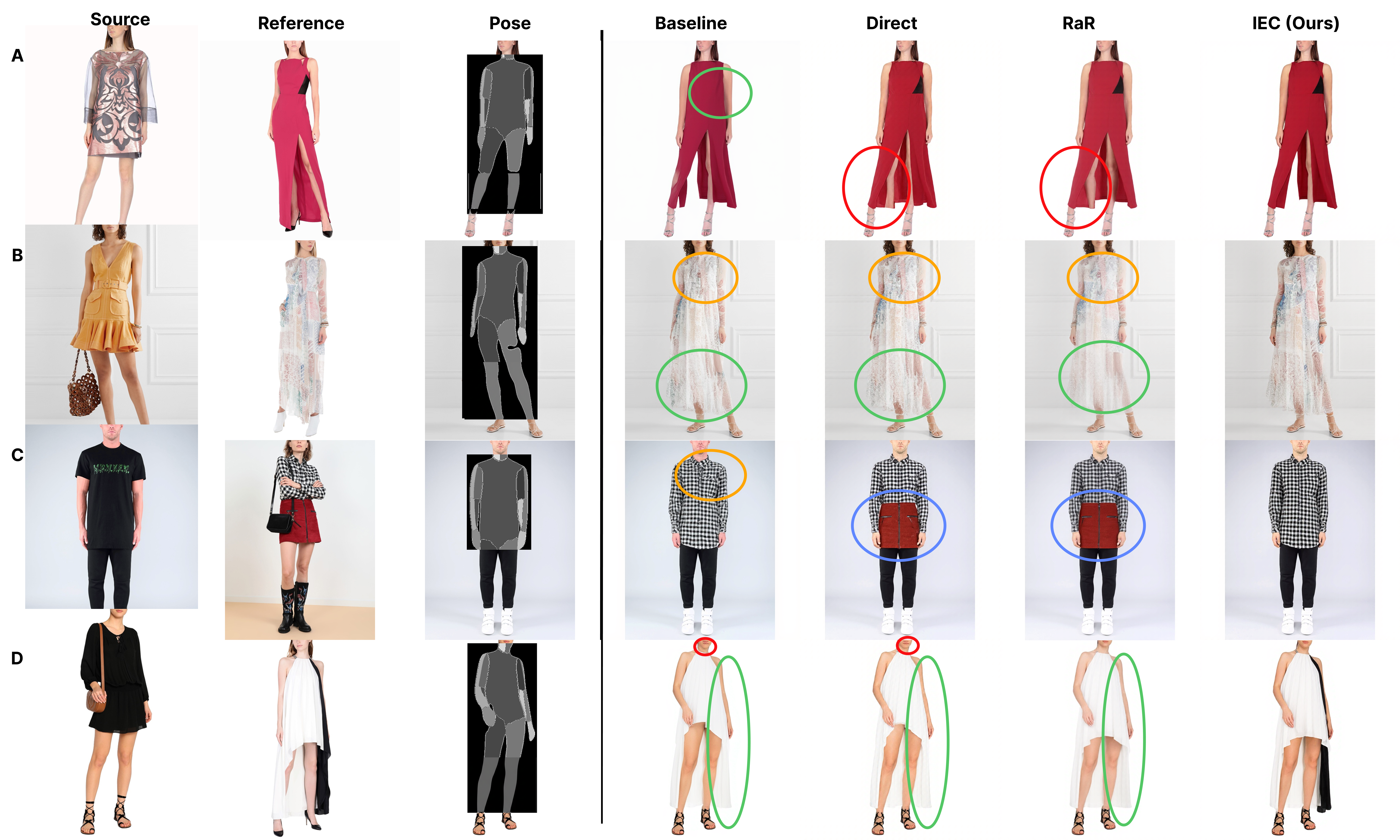}
}

\twocolumn[
  \icmltitle{When Rubrics Fail: Error Enumeration as Reward in Reference-Free RL Post-Training for Virtual Try-On}



  \icmlsetsymbol{equal}{*}

  \begin{icmlauthorlist}
    \icmlauthor{Wisdom Ikezogwo}{yyy,comp}
    \icmlauthor{Mehmet Saygin Seyfioglu}{comp}
    \icmlauthor{Ranjay Krishna}{yyy,allen}
    \icmlauthor{Karim Bouyarmane}{comp}
  \end{icmlauthorlist}

  \icmlaffiliation{yyy}{University of Washington, Seattle, USA}
  \icmlaffiliation{comp}{Amazon, Seattle, USA}
  \icmlaffiliation{allen}{Allen Institute for AI}

  \icmlcorrespondingauthor{Mehmet Saygin Seyfioglu}{mseyfiog@amazon.com }

  \icmlkeywords{VTO, ICML}

  \vskip 0.3in

  \centering
  \usebox{\teaserbox}
  \captionof{figure}{\textbf{Qualitative comparison of reward designs on non-flat references}: Given the input, IEC avoids \textcolor{lightgreen}{\emph{attribute preservation}} errors across 3 examples, \textcolor{lightblue}{\emph{garment transfer}} failures in example C, \textcolor{red}{\emph{source integrity}} issues in A and D, and \textcolor{orange}{\emph{realism}} errors in examples B and C, over all other reward designs. Per-category radar plots are in Appendix~\ref{supp:radar}.
  }
  \label{fig:teaser-main}
  
  \vskip 0.3in
  
]



\printAffiliationsAndNotice{}  

  \input{sec/0_abstract}    
  \input{sec/1_intro}
  \input{sec/3_method}

  \input{sec/4_experiments}

\input{sec/5_results_and_ablations}
  \input{sec/2_related_work}

\input{sec/6_conclusion}

\bibliography{main}
\bibliographystyle{icml2026}

\newpage
\appendix
\onecolumn

\input{sec/X_suppl}


\end{document}


  \centering
  \includegraphics[width=\textwidth, height=0.4\textheight, keepaspectratio]{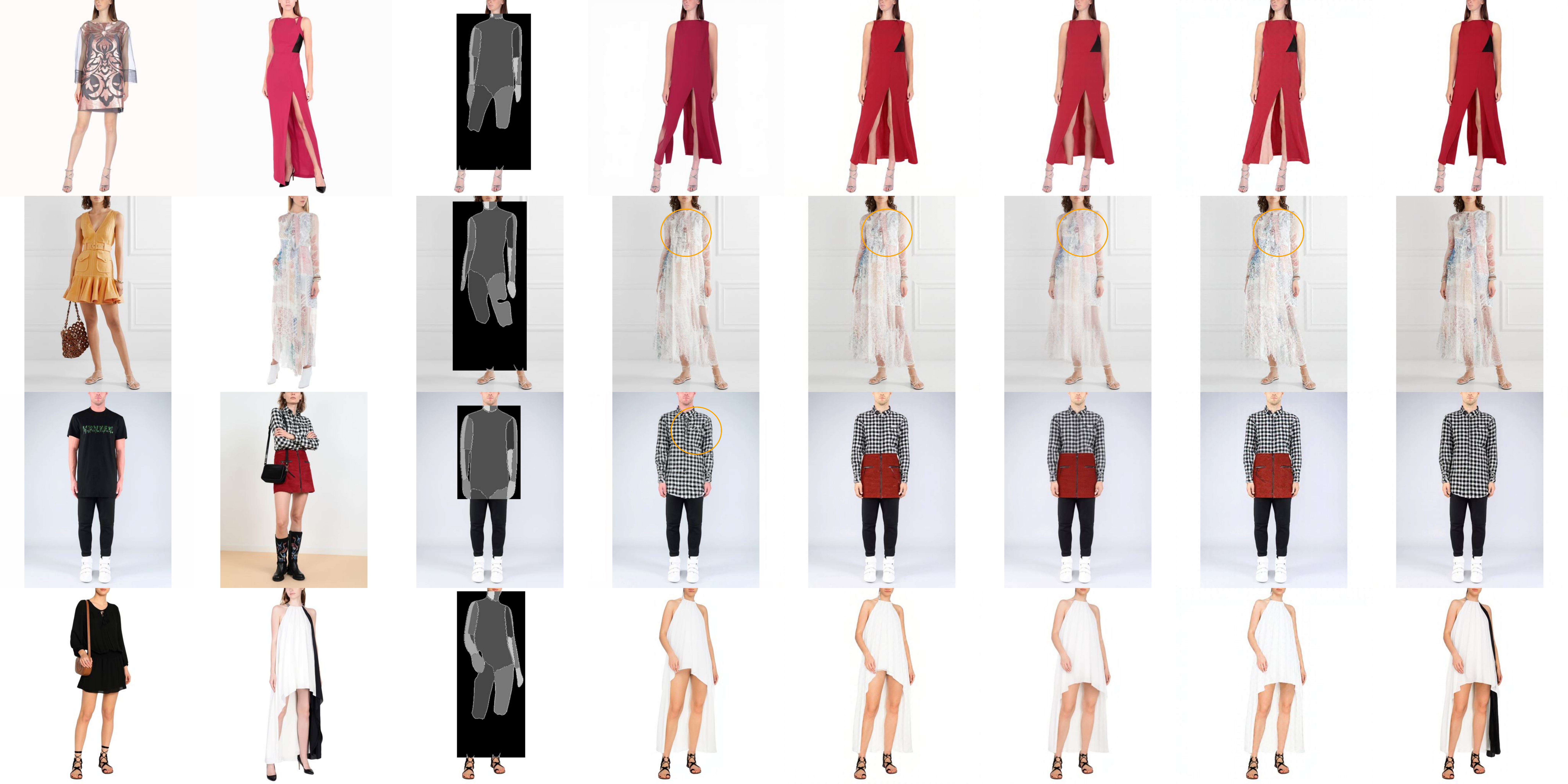}
  \captionof{figure}{Qualitative comparison of reward designs for virtual try-on post-training. Given a source image and reference garment, we show the inpainted input and outputs from the supervised baseline, Direct scoring, Rubrics-as-Rewards (RaR), and our proposed Implicit Error Counting (IEC). IEC produces the most accurate garment transfer with correct sleeve lengths, neckline shapes, and pattern preservation, while other methods exhibit subtle errors such as incorrect hem lengths (row 1, Direct), pattern distortion (row 2, RaR), or color shifts (row 3, baseline). Explicit Error Counting (EEC), an ablation variant, is analyzed in Section~\ref{sec:ablation}. \textcolor{lightblue}{same seed/noise}}
  \label{fig:teaser}

3622 - without group norm
and calibration on and best of 12 in p54 
gen 12, best of n 12 224416_oldecheckpoint-80-0
gen 24, best of n 8 224416_60-0

olde ours .

%% file: sec/0_abstract.tex
\begin{abstract}

Reinforcement learning with verifiable rewards (RLVR) and Rubrics as Rewards (RaR) have driven strong gains in domains with clear correctness signals and even in subjective domains by synthesizing evaluation criteria from ideal reference answers. But many real-world tasks admit multiple valid outputs and lack the single ideal answer that rubric generation depends on. We identify this \emph{reference-free} setting as a gap in current post-training methods and propose \textbf{Implicit Error Counting (IEC)} to fill it. Instead of checking what a response gets right against a rubric, IEC enumerates what it gets wrong, applying severity-weighted scores across task-relevant axes and converting them into calibrated per-aspect rewards. We show that na\"ive explicit enumeration is too noisy for stable optimization, and that two design choices: implicit score emission and group calibration are necessary to make error counting a reliable reward. As a case study, we validate IEC on virtual try-on (VTO), a domain that is simultaneously too constrained for holistic scoring and too permissive for rubric-based evaluation: subtle garment errors are unacceptable, yet many output variations are correct. We introduce Cascaded Error Counting (CEC) as an evaluation metric, which tracks human preferences well (60\% top-1 vs.\ 30\% others), and curate Mismatch-DressCode (MDressBench), a benchmark with maximal attribute mismatch to stress-test reward designs. On MDressBench, IEC outperforms RaR across all metrics (CEC: 5.31 vs.\ 5.60 on flat references; 5.20 vs.\ 5.53 on non-flat). On VITON-HD and DressCode, IEC matches or surpasses six baselines on 6 of 8 perceptual metrics. These results suggest that when ideal answers are unavailable, counting errors provide a stronger signal than constructing rubrics.


\end{abstract}

%% file: sec/1_intro.tex
\section{Introduction}
\label{sec:intro}

Reinforcement learning has become a standard post-training step for aligning generative models with human expectations. In domains with verifiable outcomes---mathematics, code generation, game playing---reward is straightforward: a scoring function or test environment confirms correctness~\cite{shao2024deepseekmath,ouyang2022training}. Reinforcement Learning with Verifiable Rewards (RLVR)~\cite{shao2024deepseekmath, rafailov2023direct, wallace2024diffusion,black2023training,fan2024reinforcement, xue2025dancegrpo} produces dramatic gains precisely because the reward signal is cheap, unambiguous, and dense.

Real-world tasks are rarely so clean. Medical diagnosis, scientific reasoning, and creative generation require nuanced, multi-criteria evaluation with no single correct answer. Rubrics as Rewards (RaR)~\cite{gunjal2025rubrics} addresses this gap: given a prompt and an \emph{ideal reference answer}, it synthesizes instance-specific rubric criteria and scores responses against them. RaR provides structured, interpretable supervision that outperforms coarse Likert scoring. Its core assumption, however, is that an ideal answer exists and can ground rubric generation. When it does, as in medical QA with expert-written references, rubrics capture subtle quality dimensions that binary correctness cannot.

But many domains lack ideal answers altogether. In embodied control, multiple action sequences can achieve the same goal. In open-ended design tasks, ``correctness'' is defined not by what the output matches, but by what it avoids---wrong colors, broken geometry, identity drift. We refer to this as the \textbf{reference-free} setting: the task is subjective enough that rubrics are needed, yet no ideal output exists to generate them from. In this setting, the nature of quality assessment inverts. Rather than asking ``does this response satisfy a checklist derived from an ideal answer?'' one asks ``what specific errors does this response exhibit?'' When multiple outputs are plausible, enumerating failures is more stable than enumerating successes, because the space of errors is typically smaller and more structured than the space of valid completions.

We test this hypothesis with \textbf{Implicit Error Counting (IEC)}, a reward framework for reinforcement post-training in reference-free domains. Given a candidate output and its conditioning inputs, a multimodal judge enumerates errors across task-relevant evaluation axes, weights them by severity, and emits calibrated per-aspect scores. The ``implicit'' qualifier is critical: asking the judge to expose raw error lists (Explicit Error Counting, or EEC) introduces high variance from inconsistent enumeration, where near-identical outputs receive different error counts due to surface-form variation in the judge's language. IEC instead asks the judge to \emph{internalize} the counting and output only scores, preserving the conceptual grounding of error enumeration while stabilizing the reward signal. A lightweight group calibration step further reduces prompt-to-prompt scale drift. Notably, IEC is also more efficient: it requires a single judge call per candidate, whereas RaR requires two (rubric generation followed by evaluation).

We validate IEC on virtual try-on (VTO) as a case study that exemplifies the reference-free challenge. VTO synthesizes images of users wearing target garments~\cite{choi2021viton,morelli2022dresscode,zhu2023tryondiffusion, li2025dit, seyfioglu2024diffuse}, and its failure modes are simultaneously subtle and consequential: a slightly wrong sleeve length, a discontinuous pattern, or a shifted neckline can render a result unusable. Crucially, many distinct outputs are correct; the garment can drape differently, lighting can vary, and fine details can differ. So no single ideal answer grounds rubric generation. At the same time, errors are \emph{enumerable} and \emph{localizable}: they fall along well-defined axes (garment transfer, attribute preservation, realism, lighting, source integrity) and can be detected by multimodal judges. Supervised objectives on \textit{narrow} datasets~\cite{morelli2022dresscode,choi2021viton} cannot cover the combinatorial space of poses, body shapes, textures, and garment types, and tend toward shortcut learning~\cite{labs2025flux, wu2025qwen, uniworldv2}, especially copy-paste behavior in masked-image editing. This motivates on-policy post-training that improves behavior without new paired supervision.

To stress-test reward designs under distribution shift, we curate \textbf{Mismatch-DressCode (MDressBench)}, a benchmark of 700 source-reference pairs selected for maximal attribute disagreement (e.g., short-sleeve source with long-sleeve reference). We also introduce \textbf{Cascaded Error Counting (CEC)}, an evaluation metric that builds a shared error vocabulary across candidates, enabling consistent measurement without ground truth. In a human study validating CEC as a selection metric, it achieves 60\% top-1 accuracy in recovering annotator preferences, compared to 30\% for both direct scoring and RaR-based selection.

Across MDressBench, IEC consistently outperforms both direct scoring and RaR on all metrics, achieving relative improvements of 5.96\% (CEC), 2.32\% (garment transfer), 3.48\% (attribute preservation), 4.33\% (realism), 3.3\% (lighting), and 0.6\% (source integrity) over RaR on non-flat references. On VITON-HD and DressCode, IEC outperforms all supervised and RL baselines on 6 of 8 perceptual metrics. Our ablations confirm that EEC underperforms IEC, and that group calibration provides consistent gains.



%% file: sec/3_method.tex
\section{From Rubrics to Error Counting}
\label{sec:framework}

We first formalize the gap between rubric-based and reference-free reward settings (\S\ref{sec:gap}), then present error counting as a general reward framework (\S\ref{sec:iec_general}), and finally instantiate it for virtual try-on (\S\ref{sec:vto_instantiation}).

\subsection{The Reference-Free Gap}
\label{sec:gap}

Rubrics as Rewards~\cite{gunjal2025rubrics} defines a structured reward function for each prompt $x$ via a set of $k$ rubric items $\{(w_j, c_j)\}_{j=1}^k$, where $w_j$ is the weight of criterion $j$ and $c_j : (x, \hat{y}) \mapsto \{0, 1\}$ is a binary check. The reward is:
\begin{equation}
r_{\text{rubric}}(x, \hat{y}) = \frac{\sum_{j=1}^k w_j \cdot c_j(x, \hat{y})}{\sum_{j=1}^k w_j}
\label{eq:rar}
\end{equation}
Rubric generation requires a reference answer $y^*$ to ground the criteria: each $c_j$ encodes a property that a high-quality response should satisfy, derived from analyzing $y^*$. When $y^*$ exists (e.g., expert-written medical answers), this yields precise, instance-specific supervision.

The reference-free setting arises when three conditions hold simultaneously:
\begin{enumerate}[leftmargin=1.5em,itemsep=2pt]
    \item \textbf{Multiple valid outputs exist.} Many responses $\hat{y}$ may be equally correct, so no single $y^*$ captures what ``good'' means.
    \item \textbf{Quality is defined by the absence of errors.} An output is good if it follows the conditioning and has no observable errors, not because it matches a particular reference.
    \item \textbf{Errors are enumerable and localizable.} Failure modes cluster along well-defined axes and can be detected by inspection.
\end{enumerate}
These conditions are satisfied whenever the space of acceptable outputs is large but the space of errors is structured, a property shared by visual generation, embodied control, and open-ended design tasks. Under these conditions, rubric generation from a reference answer either produces criteria that are too generic (because no single $y^*$ captures the range of valid outputs) or too specific (because $y^*$ encodes one valid solution and penalizes equally valid alternatives). Error counting exploits the structure of these domains by inverting the evaluation: instead of scoring how much an output resembles an ideal, it scores how few detectable errors it contains.

\subsection{Error Counting as a Reward Framework}
\label{sec:iec_general}

Let $x$ denote the conditioning input and $\hat{y}$ a candidate output. We define $A$ task-relevant evaluation axes $\{a_1, \ldots, a_A\}$. For each axis $a$, a judge function $f_a(x, \hat{y})$ returns a score in $[0, 1]$ by enumerating errors along that axis and mapping error counts to a monotonically decreasing score.

\paragraph{\textbf{Explicit Error Counting (EEC).}} The judge produces a list of errors $\mathcal{E}_a = \{(e_1, s_1), \ldots\}$ for each axis, where $e_i$ is an error label and $s_i \in \{\text{minor}, \text{major}\}$ is its severity. The axis score is computed as a deterministic function of the weighted count:
\begin{equation}
R_a^{\text{explicit}} = g\!\left(\sum_{(e, s) \in \mathcal{E}_a} w(s)\right)
\end{equation}
where $g$ is a monotonically decreasing mapping and $w(\cdot)$ assigns severity weights. This is conceptually clean but suffers from high variance: non-deterministic judges produce different error labels for near-identical outputs, destabilizing within-group rewards. Appendix~\ref{supp:reward_group_example} illustrates this with a concrete training group where EEC assigns a perfect score to a generation with visible errors while penalizing better candidates. See Appendix~\ref{supp:reward-error} for prompt details.

\paragraph{\textbf{Implicit Error Counting (IEC).}} We instead instruct the judge to \emph{internally} count errors for each axis and output only a calibrated score $R_a^{(i)} \in [0, 1]$, along with a short error summary for interpretability. Each error monotonically reduces the axis score; catastrophic errors exponentially reduce it toward zero. We define \emph{catastrophic} errors as wrong garment type, identity change, or a missing major garment region (see Appendix~\ref{supp:reward_hacking} / D.1). The judge implicitly determines the per-error penalty weight. The aggregate reward is $R_i^{\text{implicit}} = \frac{1}{|A|} \sum_{a=1}^{|A|} R_a^{(i)}$ \label{eq:iec}.

The key distinction from EEC is that the reward depends on the judge's \emph{internal} assessment rather than on parsed error labels, making it robust to surface-form variation of the judge, and brittleness of fixed weights \cite{gunjal2025rubrics}. The error summary provides interpretability without coupling directly to reward computation.

\paragraph{\textbf{Group Calibration.}}
\label{main:gc}
Reward magnitudes vary across prompts due to difficulty and evaluator scale, introducing unnecessary variance in RL updates. We apply group-wise calibration using robust statistics. For a group reward vector $\mathbf{R}$:
\begin{equation}
z_i = \frac{R_i - \text{median}(\mathbf{R})}{1.4826 \cdot \text{MAD}(\mathbf{R}) + \varepsilon}
\end{equation}
\begin{equation}
R_i^{\text{cal}} = \text{floor} + (1 - \text{floor}) \cdot \sigma(\alpha \cdot z_i)
\end{equation}
where MAD is the median absolute deviation, $1.4826$ rescales to standard-deviation scale under Gaussian assumptions, $\text{floor} \approx 0.05$ prevents zero rewards, and $\sigma$ is the logistic function with steepness $\alpha = 1$. This monotone transform preserves within-group ordering while reducing prompt-to-prompt scale variability.

\subsection{Instantiation: Virtual Try-On}
\label{sec:vto_instantiation}

We formulate VTO as an image-conditioned inpainting problem. Given a source image $I_s$, reference/target garment $I_r$, mask $I_m$, and pose $I_p$, the model generates $I_t$ where $I_r$ appears on $I_s$ while preserving identity, pose, and background.

\paragraph{\textbf{Architecture.}} We build on a rectified-flow backbone~\cite{liu2022flow,lipman2022flow}, following the token concatenation scheme of DiT-VTON~\cite{li2025dit}. Each image is encoded into a latent space, patchified, and concatenated: $c = x_t \circ x_r \circ x_e$, where $x_t$ is the noisy latent, $x_r$ is the reference latent, and $x_e$ is the masked source with pose stitched on. The flow-matching objective trains velocity $v_\theta(x_t, t, c)$ to approximate $v = x_1 - x_0$.

\paragraph{\textbf{GRPO Post-Training.}} We apply Group Relative Policy Optimization~\cite{shao2024deepseekmath} adapted to flow models~\cite{xue2025dancegrpo, liu2025flow}. For each condition $c$, we sample $K$ candidates and compute group-normalized advantages:
\begin{equation}
\hat{A}_i = \frac{R(x_i, c) - \text{mean}(\{R(x_j, c)\}_{j=1}^K)}{\text{std}(\{R(x_j, c)\}_{j=1}^K)}
\end{equation}

\paragraph{\textbf{Evaluation Axes.}} For VTO, we define $A=5$ axes: (i) \textcolor{lightblue}{\emph{garment transfer}} (placement, sleeve/hem length, neckline, proportions); (ii) \textcolor{lightgreen}{\emph{attribute preservation}} (color, pattern, texture); (iii) \textcolor{orange}{\emph{realism}} (drape, boundaries, halos, distortions); (iv) \textcolor{brown}{\emph{lighting consistency}}; and (v) \textcolor{red}{\emph{source integrity}} (face, hair, background unchanged). Each axis is scored by the IEC judge (Equation~\ref{eq:iec}), and the aggregate reward drives GRPO updates.

\begin{figure*}[t]
  \centering
  \includegraphics[width=\linewidth, height=0.25\textheight, keepaspectratio]{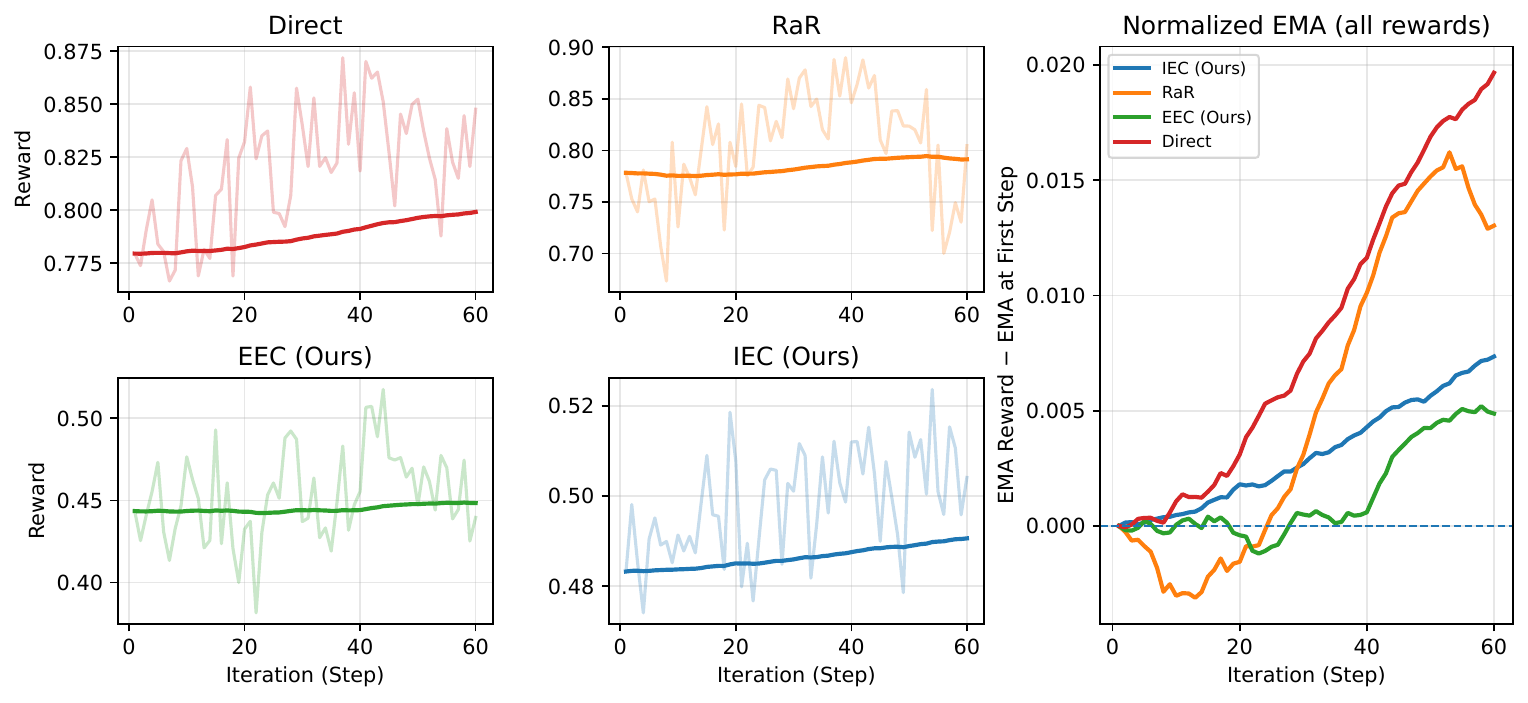}
  \caption{
    Training dynamics for four reward designs during GRPO post-training.
    \textbf{Left 2$\times$2}: Per-method curves showing raw rewards (faded) and EMA (solid, $\alpha=0.99$).
    \textbf{Right}: Normalized EMA showing relative improvement.
    IEC achieves smooth, monotonic improvement. Direct and RaR show high variance. EEC regresses early, confirming that explicit enumeration destabilizes optimization.
  }
  \label{fig:reward_dynamics}
\end{figure*}

\subsection{Baseline Rewards}
\label{sec:baselines}

We compare IEC against two standard reward designs, both evaluated by the same multimodal judge.

\paragraph{\textbf{Direct Scoring.}} The judge outputs a scalar $R_i^{\text{direct}} \in [0,1]$ for overall quality, guided by scoring bands: $[0.8, 1.0]$ for strong results, $[0.5, 0.8)$ for mixed quality, $[0.2, 0.5)$ for poor results, $[0.0, 0.2)$ for failures. It is efficient; however, it loses a fine-grained sensitivity to perceivable errors. See Appendix~\ref{supp:reward-direct}.

\paragraph{\textbf{Rubrics-as-Rewards (RaR).}} Following~\cite{gunjal2025rubrics}, we generate a prompt-specific rubric with criteria across categories \{Essential, Important, Optional, Pitfall\} with weights $w_j$. The judge evaluates each criterion with binary pass/fail. Let $\mathcal{P}_i = \{j : p_{ij} = 1, w_j > 0\}$ denote satisfied items and $\mathcal{F}_i = \{j : p_{ij} = 0, w_j < 0\}$ violated pitfalls:
\begin{equation}
R_i^{\text{rubric}} = \text{clip}\!\left(
\frac{\sum_{j \in \mathcal{P}_i} w_j + \sum_{j \in \mathcal{F}_i} w_j}{\sum_{j : w_j > 0} w_j};\, 0,\, 1
\right)
\end{equation}
This requires two judge calls (rubric generation + evaluation) and, crucially, requires generating rubrics \emph{without an ideal reference answer} in the VTO setting.

\paragraph{\textbf{Reward Cost.}} IEC and Direct scoring each require a single judge call per candidate, whereas RaR requires two (rubric generation followed by evaluation). For a group of $K{=}12$ candidates, this means IEC uses 12 judge calls per training step compared to 24 for RaR, a $2\times$ reduction in judge compute, while achieving stronger performance across all metrics.

\subsection{Cascaded Error Counting as a Metric}
\label{sec:cascade}

Plain error counting exhibits high variance as a metric because the label space drifts across images. Cascaded Error Counting reduces this by sharing an evolving error vocabulary across candidates within a group. Without pooling, a judge may enumerate zero errors for one candidate and seventeen for a near-identical one (see Appendix~\ref{supp:reward_group_example}); re-evaluating against a shared pool forces the judge to reconsider missed errors. We process candidates in blocks:
\begin{enumerate}[leftmargin=1.2em,itemsep=2pt]
\item \textbf{Pool Phase}: The judge returns preliminary per-image error sets $\mathcal{E}_{i \in B} = \{(e_i, \text{sev}), \ldots\}$. We canonicalize and merge into a pool $\mathcal{D}^{(0)}$ with semantic duplication.
\item \textbf{Verification Phase}: We re-evaluate images conditioned on $\mathcal{D}^{(0)}$, instructing the judge to find applicable pool errors and verify prior detections. This yields refined error sets $\mathcal{E}_i^{(1)}$.
\end{enumerate}
Error canonicalization applies deterministic normalization: lowercasing, removing non-alphanumeric characters, and synonym replacement. A cyclical refinement pass further stabilizes the vocabulary. Final scores are weighted counts $C_i = \sum_{(e, \text{sev}) \in \mathcal{E}_i} w(\text{sev})$ with $w(\text{major}) > w(\text{minor})$. See Appendix~\ref{supp:reward-cascade} for details.

%% file: sec/4_experiments.tex
\section{Experiments}
\label{sec:experiments}
We evaluate error counting as a reward signal using virtual try-on as our primary testbed. Below, we describe the benchmark, baselines, and evaluation protocol.

\begin{wrapfigure}{r}{0.45\columnwidth}
    \centering
    \includegraphics[width=0.4\columnwidth]{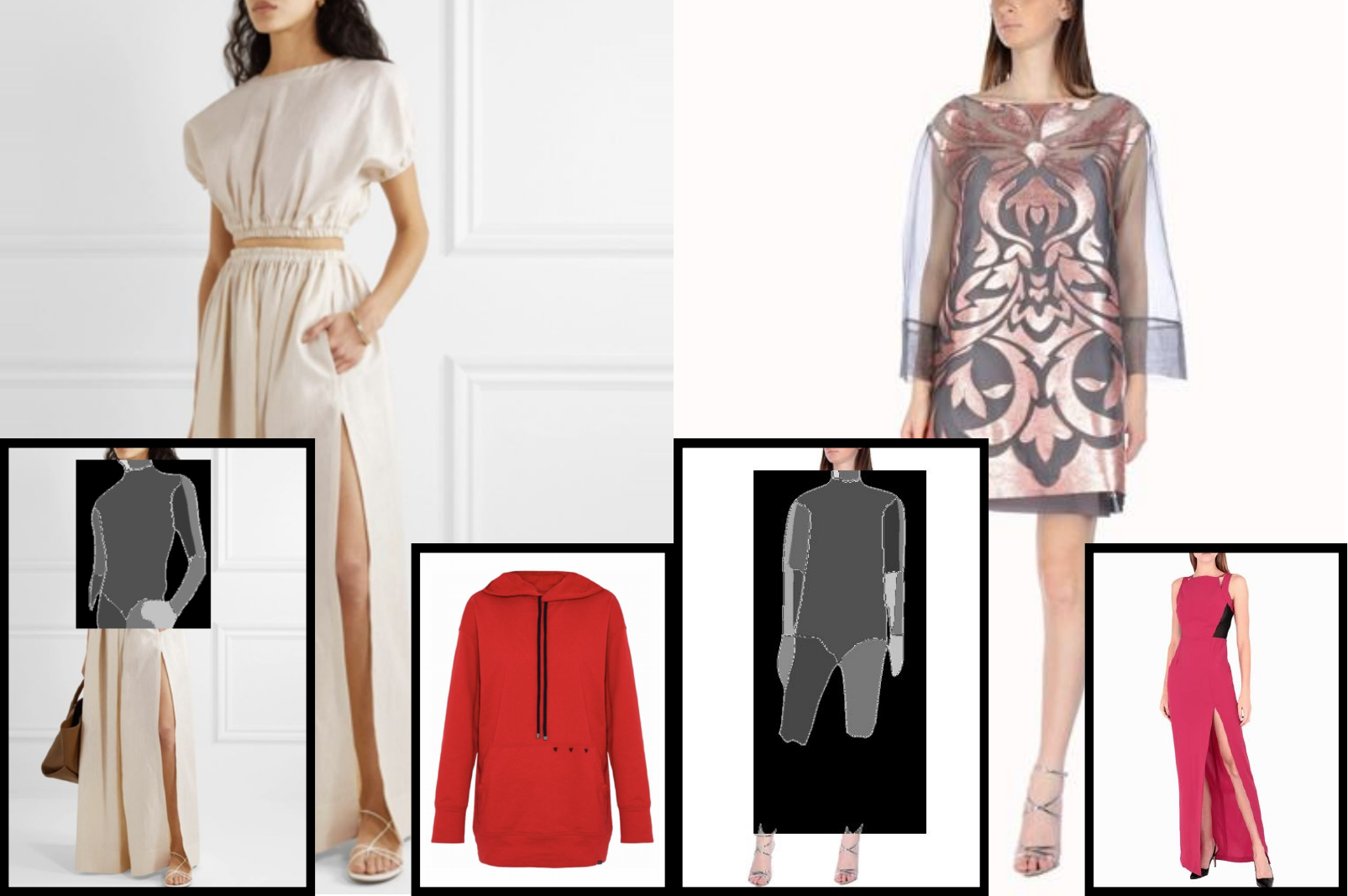}
    \caption{\textbf{MDressBench Examples.} \emph{Flat} references (\textit{left}) show the garment as a lay-flat product image on a neutral background. \emph{Non-flat} references (\textit{right}) show the garment worn by a model, requiring the model to ``extract'' the garment from arbitrary poses and backgrounds.}
    \label{fig:MDressBench}
\end{wrapfigure}

\noindent\textbf{Mismatch-DressCode (MDressBench):} Existing VTO benchmarks randomly sample source-reference pairs within categories, yielding many pairs with similar attributes that simplify the transfer task~\cite{morelli2022dresscode}. To expose the failure modes and out-of-distribution artifacts typically seen during model deployment, we curate MDressBench (Figure ~\ref{fig:MDressBench}) from DressCode (train set) by selecting pairs with \emph{maximal attribute disagreement}: short-sleeve sources with long-sleeve references, textured short dresses with ankle-length smooth references, and so on. We construct 10K pairs for RL training and reserve 350 pairs with the largest mismatch for evaluation, split across three categories (Top: 79, Lower: 122, Dress: 149) and two reference types (Flat and Non-flat), yielding 700 benchmark samples. For each pair, we follow DiT-VTON~\cite{li2025dit} to extract pose and inpainting masks for the source image, and align the reference garment crop. See Appendix~\ref{supp:mdbench} for details.

\begin{table*}[t]
\centering
\caption{Quantitative comparison on MDressBench (\textbf{flat} references). All metrics averaged across garment categories. $\uparrow$: higher is better; $\downarrow$: lower is better. Best values in \textbf{bold}. IEC outperforms all baselines across all eight metrics.}
\label{tab:main_flat}
\begingroup
\scriptsize
\setlength{\tabcolsep}{0.2pt}
\renewcommand{\arraystretch}{1.25}
\begin{tabular}{l cccccccc}
\toprule
\textbf{Method} &
\textbf{CEC}$\downarrow$ &
\textbf{RaR}$\uparrow$ &
\textbf{Direct}$\uparrow$ &
\textbf{Garm.}$\uparrow$ &
\textbf{Attr.}$\uparrow$ &
\textbf{Real.}$\uparrow$ &
\textbf{Light.}$\uparrow$ &
\textbf{Src.Int.}$\uparrow$ \\
\midrule
Baseline  & 6.018\std{1.73} & 0.760\std{0.14} & 0.796\std{0.06} & 0.793\std{0.16} & 0.815\std{0.13} & 0.735\std{0.12} & 0.760\std{0.10} & 0.914\std{0.08} \\
\midrule
Direct & 5.373\std{1.57} & 0.854\std{0.11} & 0.840\std{0.04} & 0.864\std{0.09} & 0.873\std{0.08} & 0.785\std{0.08} & 0.771\std{0.09} & 0.925\std{0.07} \\
RaR~\cite{gunjal2025rubrics} & 5.596\std{1.69} & 0.826\std{0.14} & 0.833\std{0.05} & 0.858\std{0.09} & 0.858\std{0.09} & 0.771\std{0.09} & 0.764\std{0.10} & 0.929\std{0.06} \\
\midrule
IEC (Ours) & \textbf{5.312}\std{1.72} & \textbf{0.875}\std{0.09} & \textbf{0.854}\std{0.04} & \textbf{0.873}\std{0.08} & \textbf{0.886}\std{0.07} & \textbf{0.803}\std{0.07} & \textbf{0.794}\std{0.07} & \textbf{0.933}\std{0.06} \\
\bottomrule
\end{tabular}
\endgroup
\end{table*}

\noindent\textbf{Baseline Finetuning Dataset}
We train the baseline VTO model on a data mixture of VITON-HD \cite{han2018viton, choi2021viton} and DeepFashion \footnote{https://www.kaggle.com/code/rkuo2000/deepfashion-tryon}, covering a range of poses, body shapes, viewpoints, and garment styles.

\noindent\textbf{Implementation Details}
Our model builds on Kontext-dev~\cite{labs2025flux}, a 12B rectified-flow backbone, fine-tuned for 6.2k steps on the baseline mixture using the DiT-VTON token concatenation scheme~\cite{li2025dit}. Training uses resolution $512{\times}512$, batch size 8, learning rate $5{\times}10^{-5}$, AdamW, gradient checkpointing, and bf16 precision on 8 H100 (80GB) GPUs. Complete hyperparameter settings are listed in Appendix \ref{supp:hyperparams}.

\noindent\textbf{RL Post-Training.} 
We apply GRPO with $K = 12$ generations per group and best-of-$n$ down-selection ($n = 8$) to increase reward variance. We optimize for 60 steps with batch size 3, gradient accumulation 4, and bf16 precision. Sampling uses 30 steps at $512{\times}512$ with guidance scale 2.5, so we capture fine-grained errors that show up in the latter stages of sampling, e.g., color/pattern mismatch; we use 50\% of steps for optimization biased toward earlier sampling steps. All reward computations use GPT-5-mini. Each reward design uses one judge call per candidate, except RaR (two calls: rubric generation + evaluation). To avoid self-referential bias, we use GPT-5 for evaluations.

\begin{table*}[h]
\centering
\caption{Quantitative comparison on MDressBench (\textbf{non-flat} references). All metrics averaged across garment categories. $\uparrow$: higher is better; $\downarrow$: lower is better. Best values in \textbf{bold}. IEC outperforms all baselines across all eight metrics.}
\label{tab:main_nonflat}
\begingroup
\scriptsize
\setlength{\tabcolsep}{0.2pt}
\renewcommand{\arraystretch}{1.25}
\begin{tabular}{l cccccccc}
\toprule
\textbf{Method} &
\textbf{CEC}$\downarrow$ &
\textbf{RaR}$\uparrow$ &
\textbf{Direct}$\uparrow$ &
\textbf{Garm.}$\uparrow$ &
\textbf{Attr.}$\uparrow$ &
\textbf{Real.}$\uparrow$ &
\textbf{Light.}$\uparrow$ &
\textbf{Src.Int.}$\uparrow$ \\
\midrule
Baseline & 5.489\std{1.48} & 0.811\std{0.13} & 0.815\std{0.05} & 0.835\std{0.12} & 0.848\std{0.11} & 0.754\std{0.11} & 0.767\std{0.10} & 0.902\std{0.10} \\
\midrule
Direct & 5.284\std{1.54} & 0.867\std{0.11} & 0.837\std{0.04} & 0.870\std{0.07} & 0.879\std{0.07} & 0.782\std{0.08} & 0.770\std{0.09} & 0.906\std{0.11} \\
RaR~\cite{gunjal2025rubrics} & 5.533\std{1.65} & 0.850\std{0.14} & 0.829\std{0.05} & 0.862\std{0.08} & 0.863\std{0.09} & 0.769\std{0.09} & 0.770\std{0.10} & 0.905\std{0.13} \\
\midrule

IEC w/o GC  & 5.259\std{1.51} & \textbf{0.890}\std{0.11} & 0.852\std{0.04} & 0.878\std{0.07} & 0.887\std{0.06} & 0.799\std{0.07} & \textbf{0.799}\std{0.08} & \textbf{0.918}\std{0.10} \\

IEC (Ours) & \textbf{5.203}\std{1.48} & 0.883\std{0.10} & \textbf{0.852}\std{0.04} & \textbf{0.882}\std{0.07} & \textbf{0.893}\std{0.06} & \textbf{0.803}\std{0.07} & 0.796\std{0.08} & 0.910\std{0.11} \\
\bottomrule
\end{tabular}
\endgroup
\end{table*}

\noindent\textbf{Evaluation Protocol}
For each prompt, we generate 8 outputs per model and average metrics over two runs for stability, making the effective benchmark size 5600 samples. We evaluate robustly with multiple metrics, including: (\text{i}) \textbf{CEC} ($\downarrow$) as our primary metric (\S\ref{sec:cascade}); (\text{ii}) RaR score ($\uparrow$); (\text{iii}) Direct score ($\uparrow$); and (\text{iv}) its aspect scores ($\uparrow$) for Garment Transfer, Attribute Preservation, Realism, Lighting, and Source Integrity. Additionally, we evaluate on the standard paired test splits of VITON-HD and DressCode using conventional perceptual metrics (LPIPS, SSIM, FID, KID) to assess generalization beyond MDressBench (Table~\ref{tab:standard_benchmarks}).

\noindent\textbf{Human Evaluation of CEC.} To validate that CEC reflects human preferences as a \emph{selection metric}, we conduct a 25-sample study where annotators rank try-on outputs from the baseline model across 8 images generated per sample, then evaluate each group using Direct, RaR, and CEC scoring and report top-$k$ ranking accuracy. This study validates CEC as a metric, not IEC as a training reward; a dedicated human preference study of IEC outputs at scale is left to future work. Prior work on decomposed evaluation supports the principle that counting structured violations correlates well with human quality judgments: GenEval~\cite{ghosh2023geneval} counts compositional errors in text-to-image task, and TIFA~\cite{hu2023tifa} counts question-answering failures for text-image alignment.

\noindent\textbf{Cross-Family Judge Evaluation.} Because both our training (GPT-5-mini) and evaluation (GPT-5) judges come from the same model family, we additionally evaluate, as an ablation, with a frozen open-source judge, \texttt{Qwen3.6-35B-A3B} (with temperature set to 0), using the identical CEC, RaR, and Direct pipelines, see \S\ref{sec:crossjudge_ablation}.

%% file: sec/5_results_and_ablations.tex
\section{Results}
\label{sec:results}
We evaluate on MDressBench (\emph{flat} and \emph{non-flat}). Prior to evaluation, we validate CEC as a selection metric: Table~\ref{tab:human_eval} shows CEC matches the human top choice twice as often as Direct or RaR (Top-1: 0.60 vs.\ 0.30), supporting its use as the primary evaluation signal. We organize our results around three questions: Does error counting outperform rubric-based and direct rewards? (\S\ref{sec:main_results}) Why does explicit counting fail while implicit counting succeeds? (\S\ref{sec:eec_ablation}) What role does group calibration play? (\S\ref{sec:gc_ablation})

\subsection{Error Counting Outperforms Rubrics and Direct Scoring}
\label{sec:main_results}
Tables~\ref{tab:main_flat} and~\ref{tab:main_nonflat} present results on MDressBench. All RL methods improve over the SFT baseline.
\noindent\textbf{IEC vs.\ RaR.} IEC outperforms RaR on all eight metrics under both flat and non-flat references. On flat references, IEC achieves CEC 5.31 vs.\ 5.60 for RaR. On the more challenging non-flat subset, IEC exceeds RaR by 5.96\% (CEC), 3.88\% (RaR score), 2.32\% (garment transfer), 3.48\% (attribute preservation), 4.33\% (realism), 3.3\% (lighting), and 0.6\% (source integrity). This gap is notable because RaR must generate rubrics \emph{without} an ideal reference to anchor criteria; rubrics become generic and miss prompt-specific failure modes, producing misaligned reward signals.
\noindent\textbf{IEC vs.\ Direct Scoring.} Direct scoring reduces CEC from 6.02 (SFT) to 5.37 on flat references, however, IEC still outperforms on all metrics. Direct scoring compresses quality into a single scalar, losing the fine-grained error sensitivity that IEC's aspect decomposition provides. The gap widens on non-flat references, where transfer errors are more varied and harder to capture with a direct score.
\noindent\textbf{RaR underperforms Direct.} Strikingly, RaR scores worse than Direct on CEC (5.60 vs.\ 5.37 on flat; 5.53 vs.\ 5.28 on non-flat). This confirms that in the reference-free setting, attempting to construct rubrics without ideal answers can be worse than not using rubrics at all. Binary pass/fail judgments across many criteria compound stochastic judge errors, and generic criteria miss the specific failure modes of each input.

\begin{table}[t]
\centering
\caption{Human alignment of automated selection rules ($n=25{\times}8$). CEC achieves $2\times$ higher top-1 accuracy than alternatives, validating it as a reliable selection metric.}
\label{tab:human_eval}
\setlength{\tabcolsep}{8pt}
\scriptsize
\begin{tabular}{lccc}
\toprule
\textbf{Selector} & \textbf{Top-1}$\uparrow$ & \textbf{Top-3}$\uparrow$ & \textbf{Top-5}$\uparrow$ \\
\midrule
Direct & 0.30 & 0.80 & 1.0 \\
RaR~\cite{gunjal2025rubrics} & 0.30 & 0.90 & 1.0 \\
\midrule
CEC (Ours) & \textbf{0.60} & \textbf{1.0} & 1.0 \\
\bottomrule
\end{tabular}
\end{table}

\subsection{Garment Category Analysis}
Figures~\ref{fig:category_breakdown} and~\ref{fig:teaser-main}, and Tables~\ref{tab:full_flat} and~\ref{tab:full_nonflat} show that IEC improves performance \emph{across all garment types} rather than shifting errors between categories. The largest gain is on lowers (CEC 4.66 vs.\ 5.65 for SFT), while tops are the only case where Direct is slightly better on CEC. IEC attains the best score in 20 of 24 (metric, category) combinations on flat references, indicating gains are not driven by a single category. On non-flat references, IEC wins 14/24 combinations, with gains concentrating on dresses (all 8 metrics) and tops (5/8). Figure~\ref{fig:radar_main} in Appendix~\ref{supp:radar} summarizes this per-category structure: IEC achieves the largest area across all categories, with the most pronounced gains on Dresses. See Appendix~\ref{supp:reward_hacking} for a qualitative discussion of reward hacking modes across reward designs.

\subsection{Generalization to Standard Benchmarks}
\label{sec:standard_benchmarks}

To assess whether IEC's gains transfer beyond MDressBench, we evaluate on the paired test splits of VITON-HD and DressCode using perceptual metrics. Table~\ref{tab:standard_benchmarks} compares our RL post-trained models against six external SFT baselines~\cite{catvton, catv2ton, catdm, idmvton, ootdiffusion, omnivton}. IEC achieves the best LPIPS (0.047) and SSIM (0.892) on VITON-HD, and the best FID (3.610) and ties for best SSIM (0.927) on DressCode. Across both benchmarks, IEC attains the top score on 6 of 8 metrics despite using only 60 RL steps with no additional paired data. The consistent gains over external SFT models, trained on substantially larger paired datasets, validate that error-counting rewards provide an efficient path to quality improvement.

\section{Ablations}
\label{sec:ablations}

\noindent\textbf{Why Not Explicit Error Counting?}
\label{sec:eec_ablation} A natural question is whether exposing error lists (EEC) would provide even finer-grained signal than implicit scoring. Table~\ref{tab:ablation_eec} and Figure~\ref{fig:reward_dynamics} answer this decisively: EEC \emph{fails to improve} CEC on flat references (6.03 vs.\ 6.02 SFT) and \emph{degrades} on non-flat references (5.72 vs.\ 5.49 SFT). IEC yields large improvements in both settings. This behavior is consistent with the reward dynamics in Fig.~\ref{fig:reward_dynamics}. EEC produces high-variance rewards and degrades early. Explicit label enumeration is brittle, as semantically identical errors can be counted under multiple surface forms, injecting noise into within-group comparisons and destabilizing rankings. IEC, by contrast, asks the judge to implicitly aggregate errors into a calibrated scalar, yielding more stable preference signals. More broadly, in reference-free settings, error-based rewards from multimodal VLM judges are inherently susceptible to variance from linguistic and perceptual ambiguity (see Appendix~\ref{supp:reward_group_example}). Implicit scoring is therefore not merely a convenience; it is a stability-critical design choice for optimization.
\begin{table*}[t]
\centering
\caption{Explicit vs.\ implicit error counting. EEC fails to improve over baseline; IEC achieves consistent gains.}
\label{tab:ablation_eec}
\scriptsize
\setlength{\tabcolsep}{4pt}
\renewcommand{\arraystretch}{1.15}
\begin{tabular}{lccc|ccc}
\toprule
& \multicolumn{3}{c|}{\textbf{Flat}} & \multicolumn{3}{c}{\textbf{Non-Flat}} \\
\textbf{Method} & \textbf{CEC}$\downarrow$ & \textbf{Rub.}$\uparrow$ & \textbf{Dir.}$\uparrow$ & \textbf{CEC}$\downarrow$ & \textbf{Rub.}$\uparrow$ & \textbf{Dir.}$\uparrow$ \\
\midrule
Baseline (SFT) & 6.02 & 0.76 & 0.80 & 5.49 & 0.81 & 0.82 \\
EEC & 6.03 & 0.76 & 0.81 & 5.72 & 0.77 & 0.81 \\
IEC (Ours) & \textbf{5.31} & \textbf{0.88} & \textbf{0.85} & \textbf{5.20} & \textbf{0.88} & \textbf{0.85} \\
\bottomrule
\end{tabular}
\end{table*}




\begin{figure}[h]
    \centering
    \includegraphics[width=\linewidth, height=0.25\textheight, keepaspectratio]{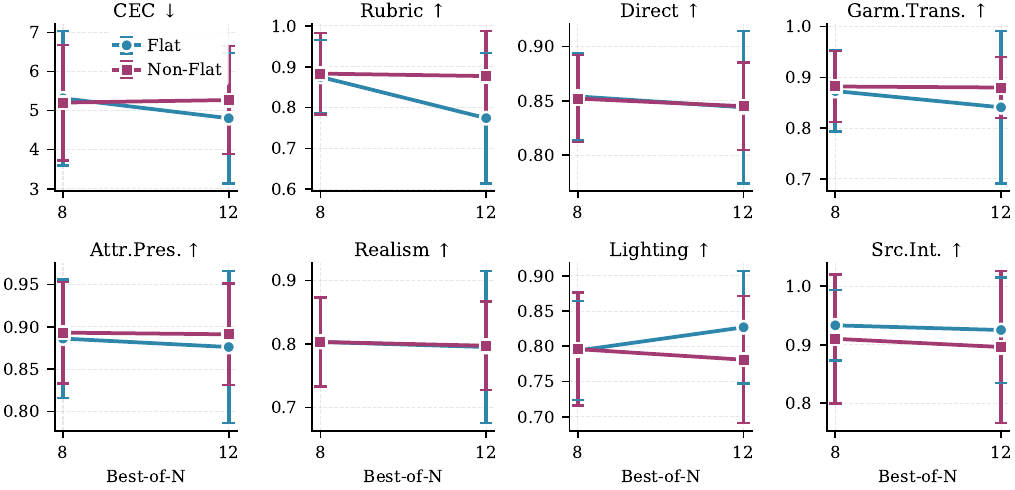}
    \caption{\textbf{Effect of Best-of-$N$} with 12 generations. Higher $N$ typically reduces reward diversity and performance.}
    \label{fig:ablation_best_of_n}
\end{figure}

\noindent\textbf{Group Calibration}
\label{sec:gc_ablation} Table~\ref{tab:main_nonflat} includes IEC w/o GC (group calibration removed). Without calibration, IEC still outperforms all baselines, confirming that the core error-counting signal is strong. However, group calibration improves CEC (5.203 vs.\ 5.259 on non-flat) by reducing prompt-to-prompt scale variability, which stabilizes GRPO's advantage estimates. The effect is more pronounced on harder prompts where reward magnitudes vary most. Note that calibration is effective for IEC precisely because the within-group ranking is reliable; for EEC, where stochastic enumeration corrupts the ranking itself (Appendix~\ref{supp:reward_group_example}), calibration does not provide equivalent benefit.

\noindent\textbf{Robustness to Underlying Judge Model}
\label{sec:crossjudge_ablation}
Because our training and evaluation judges share a model family, we ablate judge identity by evaluating the non-flat subset of the benchmark with a different and open-weight judge, as described in \S\ref{sec:experiments}. Table~\ref{tab:crossjudge} shows the relative performance across methods and judges is preserved: IEC outperforms Baseline, Direct, and RaR under \emph{both} judges on all three metrics, and the IEC$ > $Direct$ > $RaR ordering holds cross-family. This indicates IEC's gains are not an artifact of self-preference within a single judge family. Full per-category results are in Appendix~\ref{supp:b8}.

\begin{table*}[t]
\centering\scriptsize\setlength{\tabcolsep}{3pt}\renewcommand{\arraystretch}{1.1}
\caption{Judge-family ablation on MDressBench (non-flat). Evaluation using a different judge shows that IEC dominates Direct/RaR across both judges, while absolute rubric values shift under the stricter Qwen judge, the relative ordering holds.}
\label{tab:crossjudge}
\begin{tabular}{l ccc ccc}
\toprule
& \multicolumn{3}{c}{GPT-5} & \multicolumn{3}{c}{Qwen3.6-35B-A3B} \\
\cmidrule(lr){2-4}\cmidrule(lr){5-7}
Method & CEC$\downarrow$ & Rub$\uparrow$ & Dir$\uparrow$ & CEC$\downarrow$ & Rub$\uparrow$ & Dir$\uparrow$ \\
\midrule
Baseline & 5.489 & 0.811 & 0.815 & 7.114 & 0.087 & 0.879 \\
Direct   & 5.284 & 0.867 & 0.837 & 5.550 & 0.107 & 0.919 \\
RaR~\cite{gunjal2025rubrics} & 5.533 & 0.850 & 0.829 & 5.905 & 0.106 & 0.907 \\
\textbf{IEC (Ours)} & \textbf{5.203} & \textbf{0.883} & \textbf{0.852} & \textbf{5.377} & \textbf{0.110} & \textbf{0.919} \\
\bottomrule
\end{tabular}
\end{table*}

\noindent\textbf{Effect of Best-of-$N$ Sampling}
\label{sec:ablation_best_of_n}
Best-of-$N$ selects the best $N/2$ and worst $N/2$ within a group to increase diversity. We vary $N \in \{8, 12\}$ with 12 generations fixed (Figure~\ref{fig:ablation_best_of_n}). Increasing $N$ generally reduces aspect scores, consistent with a quality-diversity trade-off. CEC responds asymmetrically: it slightly worsens on non-flat references (5.20$\rightarrow$5.27) but improves on flat (5.31$\rightarrow$4.80). We use $N{=}8$ as it increases diversity within groups and is robust across both reference types.

\noindent\textbf{Qualitative Analysis}
In Figure~\ref{fig:blur}, RaR-trained models blur garment details, reflected in poor realism scores. IEC preserves fine-grained textures and patterns. Figure~\ref{fig:background_issues} shows IEC can occasionally fail on source integrity when source and reference backgrounds differ, accounting for the smaller 0.6\% improvement. This small gain is structural rather than a signal failure, because source integrity has the smallest within-group reward variance of any axis at every training window ($\sigma_{\text{src}}{=}0.060{-}0.074$ vs.\ $\sigma_{\text{garm}}{=}0.102{-}0.170$), so GRPO's group-normalized advantage on this axis is correspondingly weaker. The residual failures are predominantly pose/limb edits (38\% of labels) rather than identity/face changes (30\%). See Appendix~\ref{supp:reward_hacking} for details.

\begin{figure}[h]
    \centering
    \includegraphics[width=\linewidth, height=0.15\textheight, keepaspectratio]{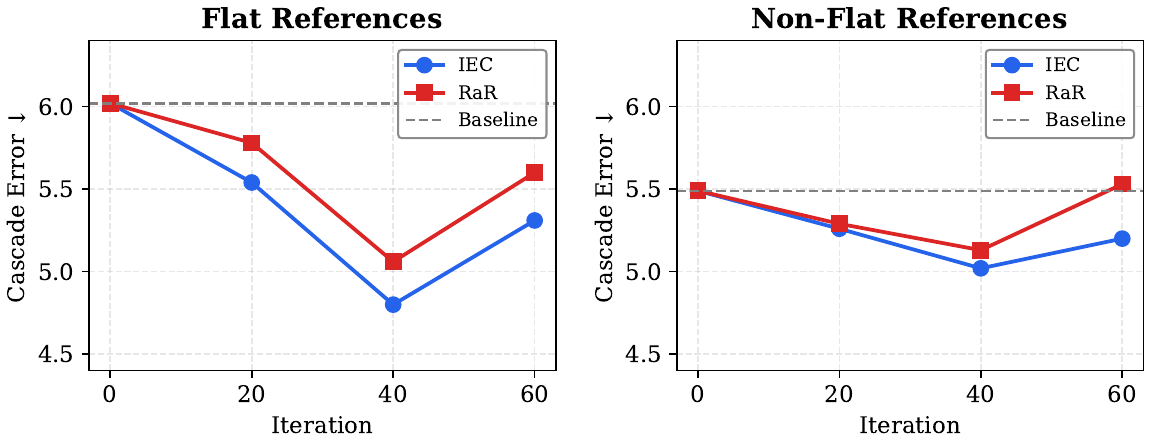}
    \caption{\textbf{Training dynamics.} CEC $\downarrow$ for IEC and RaR across post-training iterations. IEC consistently achieves lower error.}
    \label{fig:training_dynamics}
\end{figure}

\noindent\textbf{Training Dynamics}
Figure~\ref{fig:reward_dynamics} shows IEC yields smooth, monotonic reward improvement, while Direct and RaR are noisier and EEC regresses early. Figure~\ref{fig:training_dynamics} compares IEC and RaR across iterations (20, 40, 60): IEC consistently achieves lower CEC, reaching minimum of 4.80 at iteration 40 (20\% below baseline) vs.\ 5.06 for RaR. Both methods regress slightly after iteration 40, consistent with reward overfitting observed in prior RL post-training work~\cite{liu2025flow,wang2025grpoguard}. At iteration-40, with IEC a typical generation carries a tight error distribution (median 5, p90 6, max 10) with \emph{no} zero-error candidates, whereas EEC reports zero errors on 24.5\% of candidates while reaching 26 on others, direct evidence of the enumeration variance behind EEC's instability. We report at iteration-60 as RaR starts to regress; over the 0-60 window IEC's calibrated reward stays above its starting value ($\Delta_{60}{=}{+}0.018$), whereas RaR peaks around iteration-40 and is already declining by iteration-60, with its per-rubric-item fail rate climbing from 3\% to 15\% respectively. See Appendix~\ref{supp:b7} for the full per-axis analysis and training dynamics beyond iteration 60.

\begin{table*}[t]
\centering
\caption{ We evaluate SFT-only models trained on task-specific paired datasets vs.\ our RL post-training methods. Best in \textbf{bold}, second best \underline{underlined}. 
IEC matches or surpasses all external baselines on 6/8 metrics despite using only 60 RL steps with no additional paired data.}
\label{tab:standard_benchmarks}

\scriptsize
\setlength{\tabcolsep}{2pt}
\renewcommand{\arraystretch}{0.82}

\begin{tabular}{l cccc cccc}
\toprule
& \multicolumn{4}{c}{\textbf{VITON-HD}} & \multicolumn{4}{c}{\textbf{DressCode}} \\
\cmidrule(lr){2-5} \cmidrule(lr){6-9}
\textbf{Method} & LPIPS$\downarrow$ & SSIM$\uparrow$ & FID$\downarrow$ & KID$\downarrow$ & LPIPS$\downarrow$ & SSIM$\uparrow$ & FID$\downarrow$ & KID$\downarrow$ \\
\midrule
\multicolumn{9}{l}{\textit{SFT Methods}} \\
CatVTON \cite{catvton}       & 0.057 & 0.870 & 5.430 & \textbf{0.410} & 0.046 & 0.892 & 3.990 & 0.820 \\
CatVT2ON \cite{catvton}      & 0.057 & \underline{0.890} & 8.100 & 2.250 & 0.037 & \underline{0.922} & 5.720 & 2.340 \\
CAT-DM \cite{catdm}          & 0.080 & 0.877 & 5.600 & 0.830 & --    & --    & --    & --    \\
IDM-VTON \cite{idmvton}      & 0.102 & 0.870 & 6.290 & --    & 0.062 & 0.920 & 8.640 & 0.900 \\
OOTDiffusion \cite{ootdiffusion} & 0.071 & 0.878 & 8.810 & 0.820 & 0.045 & \textbf{0.927} & 4.200 & \textbf{0.370} \\
OmniVTON \cite{omnivton}     & 0.145 & 0.832 & 7.760 & --    & 0.119 & 0.865 & 5.340 & --    \\
\midrule
\multicolumn{9}{l}{\textit{Ours (SFT + RL Post-Training)}} \\
Baseline (SFT)  & 0.059 & 0.872 & 5.490 & 0.820 & 0.038 & 0.893 & 4.130 & 0.990 \\
Direct          & \underline{0.049} & 0.888 & \textbf{4.980} & \underline{0.650} & \textbf{0.029} & 0.921 & \underline{3.680} & 0.840 \\
RaR \cite{gunjal2025rubrics} & 0.055 & 0.878 & 5.210 & 0.690 & \underline{0.031} & 0.915 & 3.920 & 0.940 \\
IEC w/o GC      & 0.052 & 0.875 & 5.190 & 0.660 & 0.034 & 0.913 & 3.980 & 0.930 \\
IEC (Ours)      & \textbf{0.047} & \textbf{0.892} & \underline{5.120} & 0.660 & 0.035 & \textbf{0.927} & \textbf{3.610} & \underline{0.790} \\
\bottomrule
\end{tabular}
\end{table*}


%% file: sec/2_related_work.tex
\section{Related Work}
\label{sec:related}
\noindent\textbf{RL for Visual Generation.}
RLHF is standard for aligning language models~\cite{ouyang2022training}; DPO~\cite{rafailov2023direct} removes the explicit reward model, while GRPO~\cite{shao2024deepseekmath} normalizes advantages within candidate groups. Several works adapt policy optimization to diffusion models~\cite{black2023training,fan2024reinforcement,wallace2024diffusion}. Flow-GRPO~\cite{liu2025flow} introduces ODE-to-SDE conversion for flow-matching models; MixGRPO~\cite{li2025mixgrpo} uses mixed ODE-SDE sampling, SuperFlow~\cite{he2025superflow} enables on-the-fly training, and TreeGRPO~\cite{zhang2025treegrpo} constructs search trees for step-level credit assignment. DiffusionNFT~\cite{zheng2025diffusionnft} proposes a negative-aware finetuning objective. GRPO-Guard~\cite{wang2025grpoguard} addresses over-optimization via regulated clipping, directly relevant to the reward regression we observe (Fig.~\ref{fig:training_dynamics}). Our work focuses on reward design rather than optimization mechanics.

\noindent\textbf{Reward Design for Subjective Tasks.}
ImageReward and HPSv2~\cite{xu2024imagereward,wu2023human} train reward models on human preference datasets but are aesthetics-centric. Using VLMs as evaluators~\cite{chen2024mllm,gong2025onereward,wu2024qbench,zhang2024qbench,wang2025unified} has been applied to both evaluation and training-time rewards. RaR~\cite{gunjal2025rubrics} generates task-specific rubrics, improving interpretability; Rubicon~\cite{rubicon2025} extends this with defense rubrics, exploring explicit vs.\ implicit aggregation; and DR Tulu~\cite{shao2026drtulureinforcementlearning} evolves rubrics over training to counter generic, trivially-satisfied criteria, the same dilution we observe for reference-free rubrics. QA-LIGN~\cite{jacob2025qa} decomposes a constitution into hierarchical principle-specific QA checks scored by an LLM judge and optimized with GRPO for safety alignment using a structured per-axis reward complementary to our error-counting method. UniWorld-V2/Edit-R1~\cite{uniworldv2} uses VLM implicit feedback for image editing post-training via logit-based scoring, which is structurally parallel to our implicit approach, though applied to general editing. CoCoEdit~\cite{cocoedit2025} shows exclusive reliance on VLM rewards degrades content consistency, motivating region-based regularization---paralleling our observation that source integrity sees the smallest IEC improvement. GenEval~\cite{ghosh2023geneval} and TIFA~\cite{hu2023tifa} demonstrate that decomposing quality into countable criteria (compositional errors and QA failures, respectively) yields more stable signals than holistic scores. IEC extends this principle from evaluation to training rewards in a reference-free setting.

\noindent\textbf{Virtual Try-On and Evaluation.}
VTO synthesizes images of a person wearing a target garment while preserving identity and background. Early methods used geometric warping~\cite{han2018viton,wang2018toward}; VITON-HD~\cite{choi2021viton} introduced high-resolution try-on, DressCode~\cite{morelli2022dresscode} extended to multiple categories, LADI-VTON~\cite{morelli2023ladi} introduced latent diffusion with textual inversion, and TryOnDiffusion~\cite{zhu2023tryondiffusion} proposed parallel U-Nets. Recent methods explore DiT~\cite{peebles2023scalable} and flow-matching architectures, but rely on supervised fine-tuning on narrow datasets. VTON-VLLM~\cite{vton-vllm2025} is the closest prior work: it aligns VTO via a VLLM fine-tuned on human-validated feedback, whereas we perform on-policy GRPO post-training using error-counting rewards without human preference labels or a trained reward model. Traditional VTO metrics (FID~\cite{heusel2017gans}, LPIPS~\cite{zhang2018unreasonable}, SSIM~\cite{wang2004image}) correlate poorly with human perception of VTO-specific quality~\cite{morelli2022dresscode} and cannot detect semantic errors like incorrect sleeve length. Our CEC metric addresses this via a shared error vocabulary for consistent, fine-grained measurement.

%% file: sec/6_conclusion.tex
\section{Limitations and Broader Considerations}
\label{sec:limitations}

\textbf{Domain breadth.} We validate IEC only on VTO; testing on other reference-free domains (image editing, robotic manipulation, creative writing) would establish generality.
\textbf{Missing baselines.} We do not compare against VTON-VLLM~\cite{vton-vllm2025}, which requires a fine-tuned VLLM reward model, whereas our method is training-free.
\textbf{Metric validity.} CEC improves human alignment in our 25-sample study but remains an automated proxy; larger studies evaluating IEC outputs directly are future work.
\textbf{Reproducibility.} IEC relies on non-deterministic LLMs as judges (GPT-5-mini/GPT-5); API changes could silently alter reward signals, motivating future work with open-source judges.
\textbf{Fairness.} Our benchmark is curated from DressCode, which does not fully cover the diversity of body shapes, skin tones, and garment styles, and reward designs optimizing for judge-perceived quality may inherit judge biases. See Appendix~\ref{supp:limitations_expanded} for expanded discussion.

\section{Conclusion}
We identified the reference-free setting as a gap in RL post-training: domains where ideal answers are unavailable, multiple outputs are valid, and quality is defined by the absence of errors. We proposed Implicit Error Counting (IEC), showing that implicit score emission and group calibration are necessary for stable optimization. On VTO, IEC consistently outperforms rubric-based and direct-scoring rewards and matches or exceeds supervised baselines trained on far more paired data, at half the judge compute of RaR. When you cannot define what an ideal output looks like, define what a bad one looks like, and count.

%% file: sec/X_suppl.tex
\clearpage
\setcounter{page}{1}

\section{Mismatch-Dress Benchmark}
\label{supp:mdbench}

We construct the Mismatch-Dress Benchmark (MDressBench) through a VLM-based annotation pipeline designed to maximize attribute variation while enforcing strict visibility constraints.

\paragraph{Annotation Schema.} We define a hierarchical garment taxonomy with three primary categories (upper, lower, dress), type-specific subcategories within each (18 upper, 13 lower, 13 dress types), and a structured attribute set characterizing each garment. Required attributes include sleeve length, fit, and garment length; optional attributes include neckline, pattern, rise, waistline, and color. Each attribute takes values from a discrete enumerated set (3--8 options), enabling deterministic annotation and programmatic filtering. We additionally annotate \emph{garment presentation} $\in \{\text{masculine}, \text{feminine}, \text{unisex}\}$ independently of model gender—classification based on cut and silhouette rather than cultural convention—enabling analysis of try-on performance across gender-presentation boundaries.

\paragraph{Visibility and Extraction.} A garment is included only if fully visible: critical boundaries (hems, cuffs, necklines, waistbands) must be unoccluded and within frame, with $\geq$90\% visibility required for layered garments. Accessories are explicitly excluded. Our VLM annotation uses two-stage prompting: a system prompt defining the ontology, visibility rules, and output schema, paired with a user prompt injecting the known primary category and maximum item count. Category-specific extraction rules prevent ambiguity: upper/lower images ($n_{\max}=2$) include the ground truth category and may include the complementary category if fully visible, with coordinated outfits (e.g., shirt + skirt) split into components rather than mislabeled as dresses; dress images ($n_{\max}=1$) enforce single-piece annotation.

\paragraph{Attribute Diversity.} Let $\mathcal{S}$ denote the space of attribute combinations and $\hat{P}(s)$ the empirical frequency of combination $s$ after initial annotation. We identify underrepresented strata as:
\begin{equation}
\mathcal{S}_{\text{rare}} = \left\{ s \in \mathcal{S} : \hat{P}(s) < \frac{0.5}{|\mathcal{S}|} \right\}
\end{equation}
and actively source additional images to fill these gaps. This stratified resampling yields a balanced distribution satisfying $\max_s P(s) / \min_s P(s) \leq 5$. Allowing complementary garments in upper/lower images further increases attribute diversity while maintaining ground truth integrity. The resulting benchmark contains 700 challenging pairs spanning 44 garment types, achieving $\geq$60\% coverage of theoretically possible attribute combinations.


\section{Training and Results}

\begin{table}[h]
\centering
\caption{Training hyperparameters for SFT and RL stages.}
\label{tab:hyperparams}
\footnotesize
\setlength{\tabcolsep}{6pt}
\renewcommand{\arraystretch}{1.1}
\begin{tabular}{lcc}
\toprule
\textbf{Hyperparameter} & \textbf{SFT} & \textbf{RL} \\
\midrule
\multicolumn{3}{l}{\textit{Optimization}} \\
Batch size & 4 & 3 \\
Gradient accumulation steps & 12 & 4 \\
Learning rate & 5e-5 & 1e-5 \\
Weight decay & 1e-4 & 1e-4 \\
Max gradient norm & 1.0 & -- \\
LR warmup steps & 0 & -- \\
Mixed precision & bf16 & bf16 \\
\midrule
\multicolumn{3}{l}{\textit{RL-specific}} \\
Num generations ($K$) & -- & 12 \\
Best-of-$n$ down-selection & -- & 8 \\
Sampling steps & -- & 30 \\
Resolution (H $\times$ W) & 512 $\times$ 512  & 512 $\times$ 512 \\
CFG / guidance scale & -- & 2.5 \\
Shift & -- & 3 \\
Init same noise & -- & \checkmark \\
Timestep fraction & -- & 0.3 \\
$\eta$ (EMA) & -- & 0.5 \\
Clip range & -- & 0.2 \\
Advantage clip max & -- & 5.0 \\
KL weight & -- & 0.1 \\
Judge model & -- & GPT-5-mini \\
\bottomrule
\end{tabular}
\end{table}

\subsection{Training Hyperparameters}
\label{supp:hyperparams}

Table~\ref{tab:hyperparams} summarizes the hyperparameters used for supervised fine-tuning (SFT) and reinforcement learning (RL) stages.



\subsection{Results across MDressBench Garment Categories}
\label{app:full_results}
\begin{figure*}[ht!]
  \centering
  \includegraphics[width=\linewidth, height=0.2\textheight, keepaspectratio]{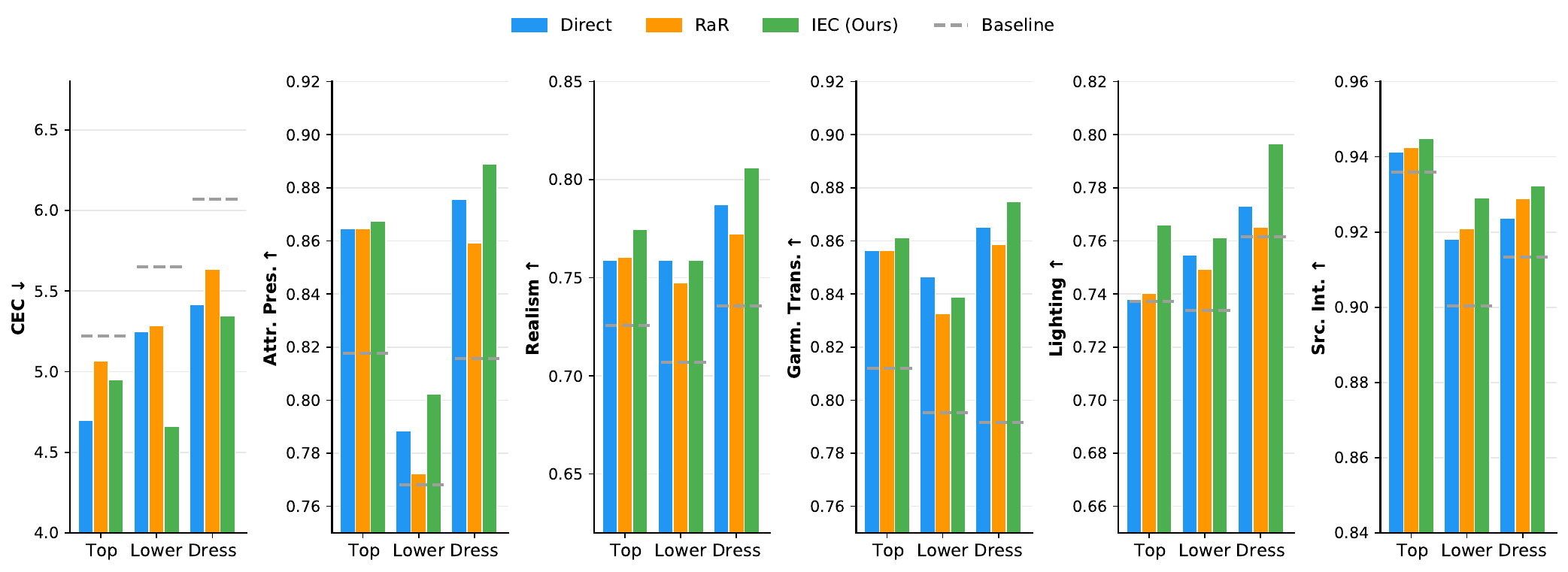}
  
  \vspace{0.5em} 
  
  \includegraphics[width=\linewidth, height=0.2\textheight, keepaspectratio]{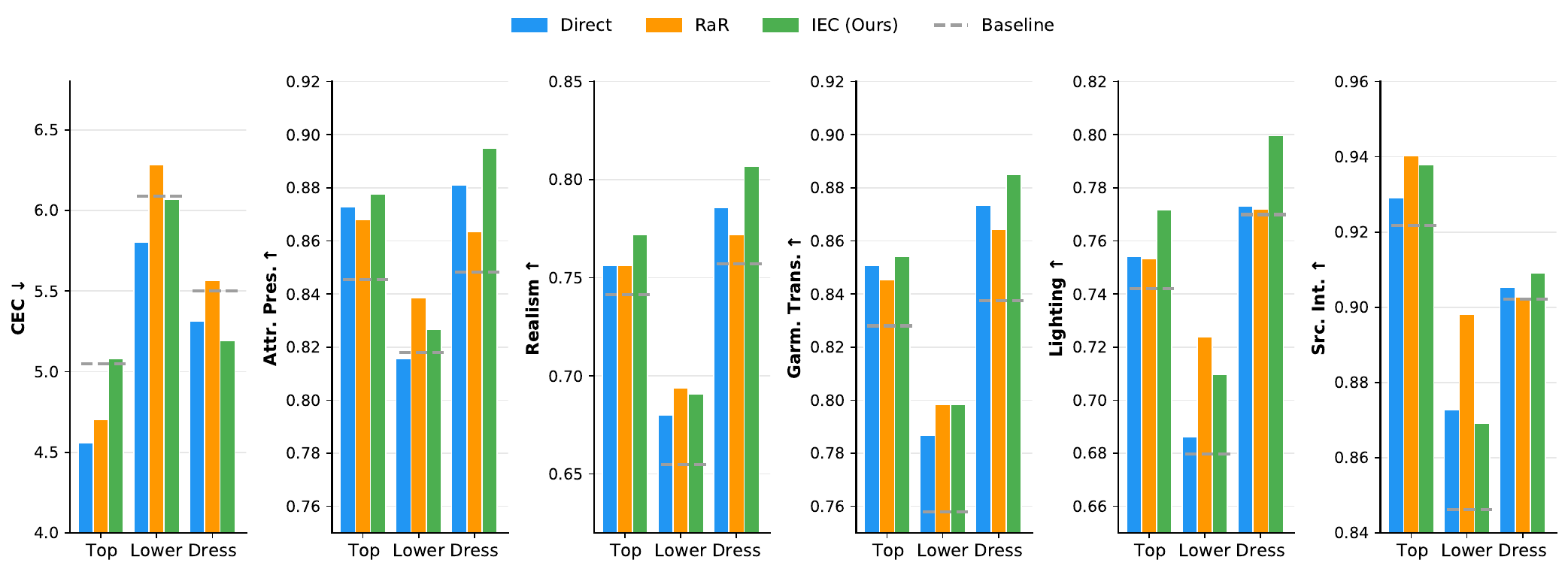}
  \caption{Garment-category breakdown on MDressBench for flat (top) and non-flat (bottom) references. IEC (green) consistently outperforms baselines across all categories. 
  }
  \label{fig:category_breakdown}
\end{figure*}

Tables~\ref{tab:full_flat}, \ref{tab:full_nonflat} and Figure~\ref{fig:category_breakdown} provide complete per-category breakdowns for all metrics on flat and non-flat reference images respectively.

\begin{table*}[ht!]
\centering
\caption{Full per-category results for flat reference images. Best values per (metric, category) in \textbf{bold}.}
\label{tab:full_flat}
\begingroup
\scriptsize
\setlength{\tabcolsep}{0.2pt}
\renewcommand{\arraystretch}{1.2}
\begin{tabular}{llcccccccc}
\toprule
\textbf{Model} & \textbf{Cat} & 
\textbf{CEC}$\downarrow$ &
\textbf{RaR}$\uparrow$ &
\textbf{Direct}$\uparrow$ &
\textbf{Garm.}$\uparrow$ &
\textbf{Attr.}$\uparrow$ &
\textbf{Real.}$\uparrow$ &
\textbf{Light.}$\uparrow$ &
\textbf{Src.Int.}$\uparrow$ \\

\midrule
Baseline.  & all & 6.018\std{1.73} & 0.760\std{0.14} & 0.796\std{0.06} & 0.793\std{0.16} & 0.815\std{0.13} & 0.735\std{0.12} & 0.760\std{0.10} & 0.914\std{0.08} \\
 & top & 5.220\std{1.15} & 0.753\std{0.14} & 0.801\std{0.03} & 0.812\std{0.10} & 0.818\std{0.13} & 0.726\std{0.10} & 0.737\std{0.09} & 0.936\std{0.03} \\
 & lower & 5.652\std{1.25} & 0.777\std{0.10} & 0.776\std{0.03} & 0.795\std{0.12} & 0.768\std{0.14} & 0.707\std{0.12} & 0.734\std{0.11} & 0.900\std{0.10} \\
 & dress & 6.072\std{1.76} & 0.760\std{0.14} & 0.796\std{0.06} & 0.792\std{0.17} & 0.816\std{0.13} & 0.736\std{0.12} & 0.762\std{0.10} & 0.913\std{0.08} \\
\midrule
Direct & all & 5.373\std{1.57} & 0.854\std{0.11} & 0.840\std{0.04} & 0.864\std{0.09} & 0.873\std{0.08} & 0.785\std{0.08} & 0.771\std{0.09} & 0.925\std{0.07} \\
 & top & \textbf{4.701}\std{1.05} & \textbf{0.833}\std{0.09} & 0.828\std{0.03} & 0.856\std{0.07} & 0.865\std{0.07} & 0.759\std{0.08} & 0.738\std{0.09} & 0.941\std{0.03} \\
 & lower & 5.250\std{1.80} & 0.856\std{0.06} & 0.809\std{0.03} & \textbf{0.847}\std{0.06} & 0.788\std{0.15} & \textbf{0.759}\std{0.06} & 0.755\std{0.08} & 0.918\std{0.07} \\
 & dress & 5.415\std{1.59} & 0.855\std{0.11} & 0.842\std{0.05} & 0.865\std{0.10} & 0.876\std{0.08} & 0.787\std{0.08} & 0.773\std{0.09} & 0.924\std{0.07} \\
\midrule
RaR & all & 5.596\std{1.69} & 0.826\std{0.14} & 0.833\std{0.05} & 0.858\std{0.09} & 0.858\std{0.09} & 0.771\std{0.09} & 0.764\std{0.10} & 0.929\std{0.06} \\
 & top & 5.066\std{1.07} & 0.805\std{0.11} & 0.829\std{0.03} & 0.856\std{0.07} & 0.865\std{0.07} & 0.761\std{0.07} & 0.740\std{0.09} & 0.943\std{0.03} \\
 & lower & 5.286\std{1.22} & 0.824\std{0.07} & 0.799\std{0.04} & 0.833\std{0.08} & 0.772\std{0.14} & 0.747\std{0.08} & 0.749\std{0.10} & 0.921\std{0.09} \\
 & dress & 5.634\std{1.72} & 0.827\std{0.14} & 0.834\std{0.05} & 0.859\std{0.09} & 0.859\std{0.09} & 0.772\std{0.09} & 0.765\std{0.10} & 0.929\std{0.06} \\
\midrule
EEC & all & 6.032\std{1.86} & 0.761\std{0.19} & 0.807\std{0.06} & 0.846\std{0.10} & 0.849\std{0.09} & 0.735\std{0.10} & 0.709\std{0.11} & 0.915\std{0.06} \\
 & top & 5.355\std{1.01} & 0.714\std{0.15} & 0.797\std{0.03} & 0.834\std{0.07} & 0.848\std{0.07} & 0.711\std{0.08} & 0.677\std{0.10} & 0.935\std{0.03} \\
 & lower & 5.830\std{1.58} & 0.762\std{0.11} & 0.776\std{0.03} & 0.821\std{0.08} & 0.775\std{0.14} & 0.710\std{0.08} & 0.696\std{0.09} & 0.903\std{0.07} \\
 & dress & 6.076\std{1.90} & 0.764\std{0.19} & 0.808\std{0.06} & 0.847\std{0.10} & 0.851\std{0.09} & 0.737\std{0.10} & 0.711\std{0.11} & 0.914\std{0.06} \\
\midrule
IEC & all & \textbf{5.312}\std{1.72} & \textbf{0.875}\std{0.09} & \textbf{0.854}\std{0.04} & \textbf{0.873}\std{0.08} & \textbf{0.886}\std{0.07} & \textbf{0.803}\std{0.07} & \textbf{0.794}\std{0.07} & \textbf{0.933}\std{0.06} \\
 & top & 4.951\std{1.25} & 0.819\std{0.12} & \textbf{0.838}\std{0.03} & \textbf{0.861}\std{0.06} & \textbf{0.867}\std{0.08} & \textbf{0.774}\std{0.07} & \textbf{0.766}\std{0.08} & \textbf{0.945}\std{0.03} \\
 & lower & \textbf{4.661}\std{1.79} & \textbf{0.872}\std{0.04} & \textbf{0.813}\std{0.03} & 0.839\std{0.07} & \textbf{0.802}\std{0.15} & \textbf{0.759}\std{0.07} & \textbf{0.761}\std{0.08} & \textbf{0.929}\std{0.05} \\
 & dress & \textbf{5.347}\std{1.74} & \textbf{0.878}\std{0.09} & \textbf{0.856}\std{0.04} & \textbf{0.875}\std{0.09} & \textbf{0.889}\std{0.07} & \textbf{0.806}\std{0.07} & \textbf{0.796}\std{0.07} & \textbf{0.932}\std{0.06} \\
\bottomrule
\end{tabular}
\endgroup
\end{table*}

\begin{table*}[ht!]
\centering
\caption{Full per-category results for non-flat reference images. Best values per (metric, category) in \textbf{bold}.}
\label{tab:full_nonflat}
\begingroup
\scriptsize
\setlength{\tabcolsep}{0.2pt}
\renewcommand{\arraystretch}{1.2}
\begin{tabular}{llcccccccc}
\toprule
\textbf{Model} & \textbf{Cat} & 
\textbf{CEC}$\downarrow$ &
\textbf{RaR}$\uparrow$ &
\textbf{Direct}$\uparrow$ &
\textbf{Garm.}$\uparrow$ &
\textbf{Attr.}$\uparrow$ &
\textbf{Real.}$\uparrow$ &
\textbf{Light.}$\uparrow$ &
\textbf{Src.Int.}$\uparrow$ \\
\midrule
Baseline. & all & 5.489\std{1.48} & 0.811\std{0.13} & 0.815\std{0.05} & 0.835\std{0.12} & 0.848\std{0.11} & 0.754\std{0.11} & 0.767\std{0.10} & 0.902\std{0.10} \\
 & top & 5.049\std{1.03} & 0.837\std{0.07} & 0.808\std{0.03} & 0.828\std{0.09} & 0.845\std{0.10} & 0.741\std{0.09} & 0.742\std{0.10} & 0.922\std{0.08} \\
 & lower & 6.089\std{1.98} & 0.759\std{0.18} & 0.743\std{0.08} & 0.758\std{0.13} & 0.818\std{0.11} & 0.655\std{0.14} & 0.680\std{0.15} & 0.846\std{0.15} \\
 & dress & 5.502\std{1.48} & 0.811\std{0.13} & 0.817\std{0.05} & 0.837\std{0.12} & 0.848\std{0.11} & 0.757\std{0.11} & 0.770\std{0.10} & 0.902\std{0.10} \\
\midrule
Direct & all & 5.284\std{1.54} & 0.867\std{0.11} & 0.837\std{0.04} & 0.870\std{0.07} & 0.879\std{0.07} & 0.782\std{0.08} & 0.770\std{0.09} & 0.906\std{0.11} \\
 & top & \textbf{4.559}\std{0.99} & \textbf{0.846}\std{0.06} & 0.827\std{0.03} & 0.851\std{0.09} & 0.873\std{0.07} & 0.756\std{0.09} & 0.754\std{0.08} & 0.929\std{0.07} \\
 & lower & 5.804\std{1.14} & 0.848\std{0.13} & 0.761\std{0.08} & 0.787\std{0.10} & 0.816\std{0.12} & 0.680\std{0.11} & 0.686\std{0.11} & 0.873\std{0.14} \\
 & dress & 5.316\std{1.56} & 0.868\std{0.11} & 0.840\std{0.04} & 0.873\std{0.07} & 0.881\std{0.07} & 0.786\std{0.08} & 0.773\std{0.09} & 0.905\std{0.11} \\
\midrule
RaR & all & 5.533\std{1.65} & 0.850\std{0.14} & 0.829\std{0.05} & 0.862\std{0.08} & 0.863\std{0.09} & 0.769\std{0.09} & 0.770\std{0.10} & 0.905\std{0.13} \\
 & top & 4.704\std{1.27} & 0.838\std{0.12} & 0.827\std{0.04} & 0.845\std{0.09} & 0.868\std{0.07} & 0.756\std{0.09} & 0.753\std{0.09} & \textbf{0.940}\std{0.04} \\
 & lower & 6.286\std{1.56} & \textbf{0.852}\std{0.13} & \textbf{0.783}\std{0.08} & \textbf{0.798}\std{0.11} & \textbf{0.838}\std{0.09} & \textbf{0.694}\std{0.12} & \textbf{0.724}\std{0.10} & \textbf{0.898}\std{0.11} \\
 & dress & 5.565\std{1.65} & 0.851\std{0.14} & 0.830\std{0.05} & 0.864\std{0.08} & 0.863\std{0.09} & 0.772\std{0.09} & 0.772\std{0.10} & 0.903\std{0.13} \\
\midrule
EEC & all & 5.721\std{1.70} & 0.769\std{0.18} & 0.806\std{0.05} & 0.856\std{0.08} & 0.855\std{0.08} & 0.739\std{0.09} & 0.712\std{0.11} & 0.893\std{0.12} \\
 & top & 4.980\std{1.32} & 0.763\std{0.19} & 0.802\std{0.04} & 0.835\std{0.08} & 0.852\std{0.07} & 0.725\std{0.09} & 0.697\std{0.10} & 0.926\std{0.06} \\
 & lower & \textbf{5.411}\std{1.30} & 0.748\std{0.11} & 0.755\std{0.05} & 0.787\std{0.10} & 0.801\std{0.11} & 0.667\std{0.10} & 0.661\std{0.11} & 0.885\std{0.10} \\
 & dress & 5.771\std{1.72} & 0.770\std{0.18} & 0.808\std{0.05} & 0.859\std{0.07} & 0.857\std{0.08} & 0.741\std{0.09} & 0.714\std{0.11} & 0.891\std{0.12} \\
\midrule
IEC & all & \textbf{5.203}\std{1.48} & \textbf{0.883}\std{0.10} & \textbf{0.852}\std{0.04} & \textbf{0.882}\std{0.07} & \textbf{0.893}\std{0.06} & \textbf{0.803}\std{0.07} & \textbf{0.796}\std{0.08} & \textbf{0.910}\std{0.11} \\
 & top & 5.079\std{1.25} & 0.826\std{0.08} & \textbf{0.837}\std{0.03} & \textbf{0.854}\std{0.09} & \textbf{0.878}\std{0.08} & \textbf{0.772}\std{0.07} & \textbf{0.772}\std{0.08} & 0.938\std{0.05} \\
 & lower & 6.071\std{1.57} & 0.841\std{0.15} & 0.773\std{0.08} & \textbf{0.798}\std{0.10} & 0.827\std{0.10} & 0.691\std{0.11} & 0.710\std{0.11} & 0.869\std{0.15} \\
 & dress & \textbf{5.191}\std{1.48} & \textbf{0.888}\std{0.09} & \textbf{0.855}\std{0.04} & \textbf{0.885}\std{0.07} & \textbf{0.895}\std{0.06} & \textbf{0.807}\std{0.07} & \textbf{0.800}\std{0.08} & \textbf{0.909}\std{0.11} \\
\bottomrule
\end{tabular}
\endgroup
\end{table*}


\subsection{Garment-Category Radar Plot}
\label{supp:radar}

\begin{figure}[t]
  \centering
  \includegraphics[width=0.6\linewidth, keepaspectratio]{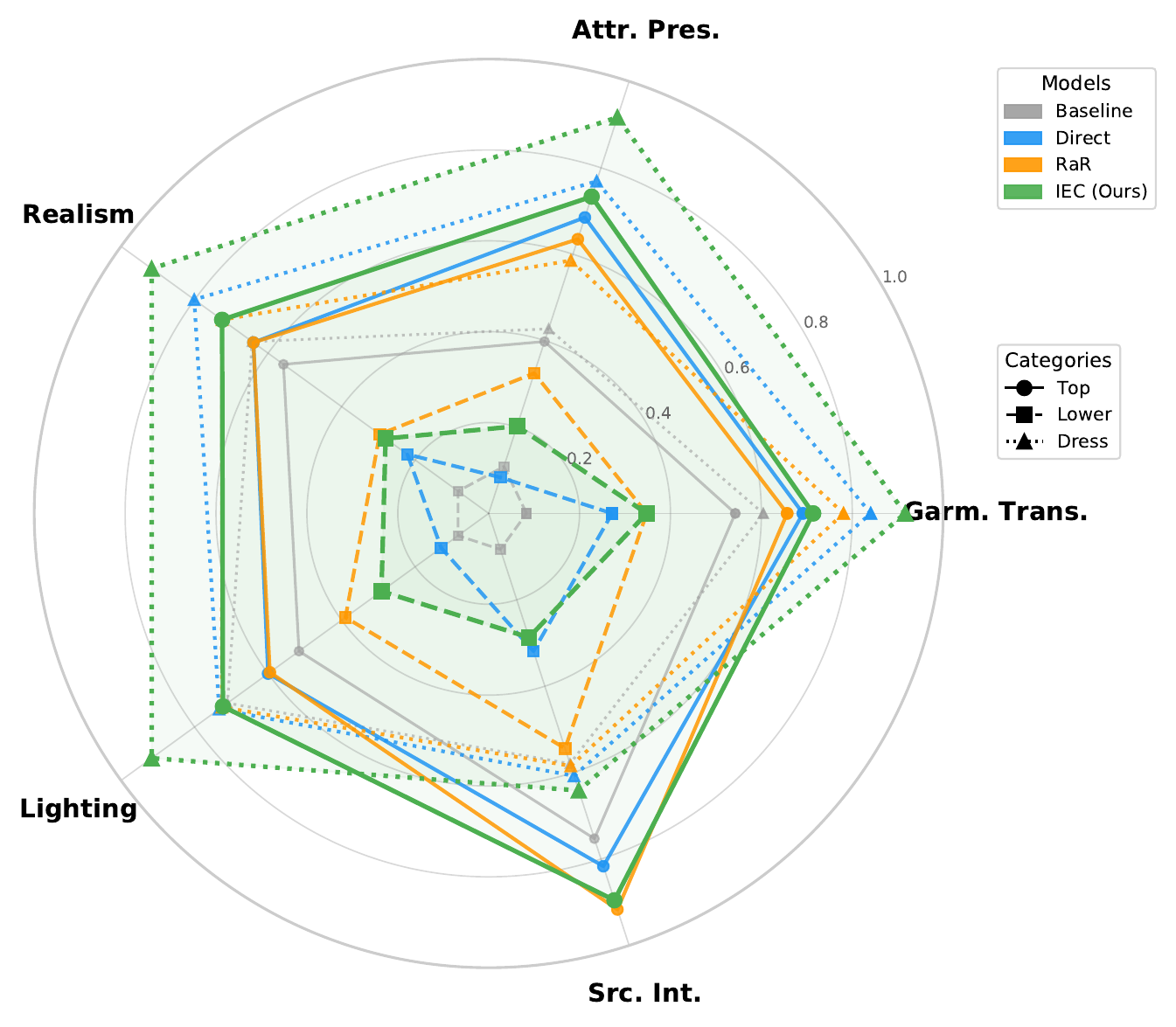}
  \caption{\textbf{Garment-category quality metrics} on \textbf{non-flat} references. IEC (green) consistently achieves the largest area across all categories, with the most gains on Dresses.}
  \label{fig:radar_main}
\end{figure}

Figure~\ref{fig:radar_main} shows per-category quality metrics on non-flat references. IEC consistently achieves the largest area across all categories, with the most pronounced gains on Dresses.

\begin{figure}[h]
  \centering
  \includegraphics[width=\linewidth, height=0.25\textheight, keepaspectratio]{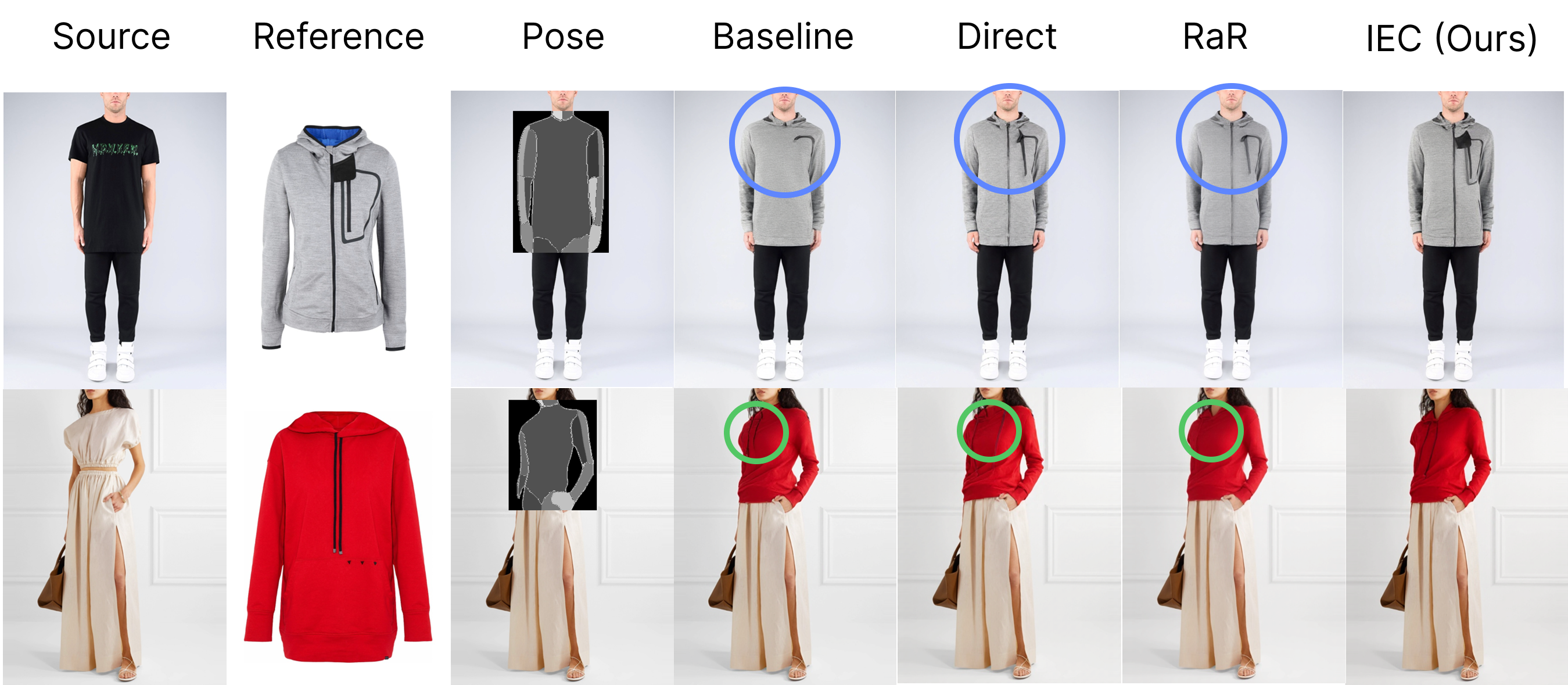}
  \caption{\textbf{Qualitative comparison of reward designs on flat references}: Given the pose-stitched masked source and the flat reference, IEC avoids \textcolor{lightgreen}{\emph{attribute preservation}} errors and \textcolor{lightblue}{\emph{garment transfer}} failures over all other reward designs.}
  \label{fig:flat_teaser_supp}
\end{figure}




\subsection{Expanded Limitations}
\label{supp:limitations_expanded}


\paragraph{Evaluation axes.} We use five manually defined axes for VTO. Automating axis discovery for new domains---perhaps by having the judge itself propose relevant error categories---is an open direction.

\paragraph{Dynamic error weighting.} We use equal axis weights throughout training. A curriculum that emphasizes structural correctness early and shifts toward fine-grained quality later may improve sample efficiency, analogous to curriculum-based rubric weighting suggested by~\cite{gunjal2025rubrics}.

\paragraph{Benchmark coverage.} MDressBench intentionally concentrates on hard, attribute-mismatched pairs and is curated from DressCode only, which does not cover all body shapes and garment styles. We focus on garments and ignore accessories like bags.

\subsection{Reward Hacking Modes}
\label{supp:reward_hacking}

\begin{figure*}[ht]
  \centering
  \includegraphics[width=\linewidth, height=0.3\textheight, keepaspectratio]{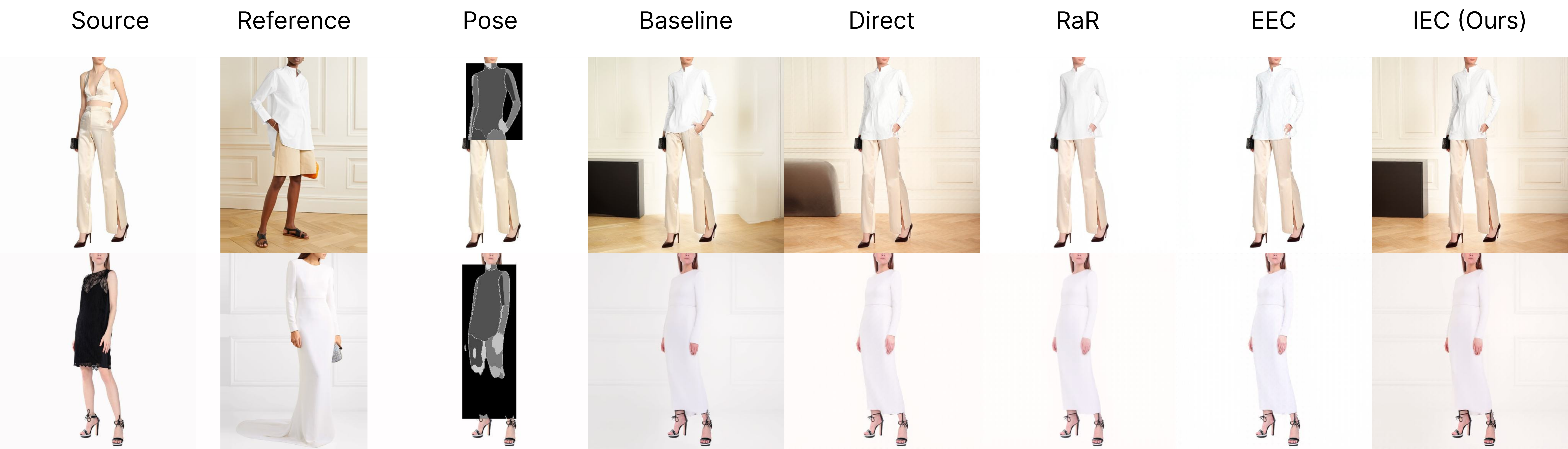}
  \caption{\textbf{Qualitative analysis}: Post-training with RaR tends to blur out garment details as shown in these examples. The garment patterns in both dresses are maintained by IEC but visually faded with RaR.
  }
  \label{fig:blur}
\end{figure*}

Reward hacking manifests differently across reward designs. Based on qualitative inspection of training outputs:

\paragraph{RaR.} The most visible hacking mode is \emph{texture smoothing}: RaR-trained models tend to blur garment details (Figure~\ref{fig:blur}), likely because the binary pass/fail rubric criteria are easier to satisfy with smooth outputs that avoid fine-grained pattern errors. This manifests as washed-out textures and reduced pattern fidelity.

\paragraph{Direct.} Direct scoring tends to produce outputs with increased color saturation and contrast, visually striking images that score well on a holistic quality scale but may not faithfully reproduce the reference garment's actual appearance. Figure~\ref{fig:flat_teaser_supp} shows qualitative comparisons on flat references of this issue.

\begin{figure*}[ht]
  \centering
  \includegraphics[width=\linewidth, height=0.3\textheight, keepaspectratio]{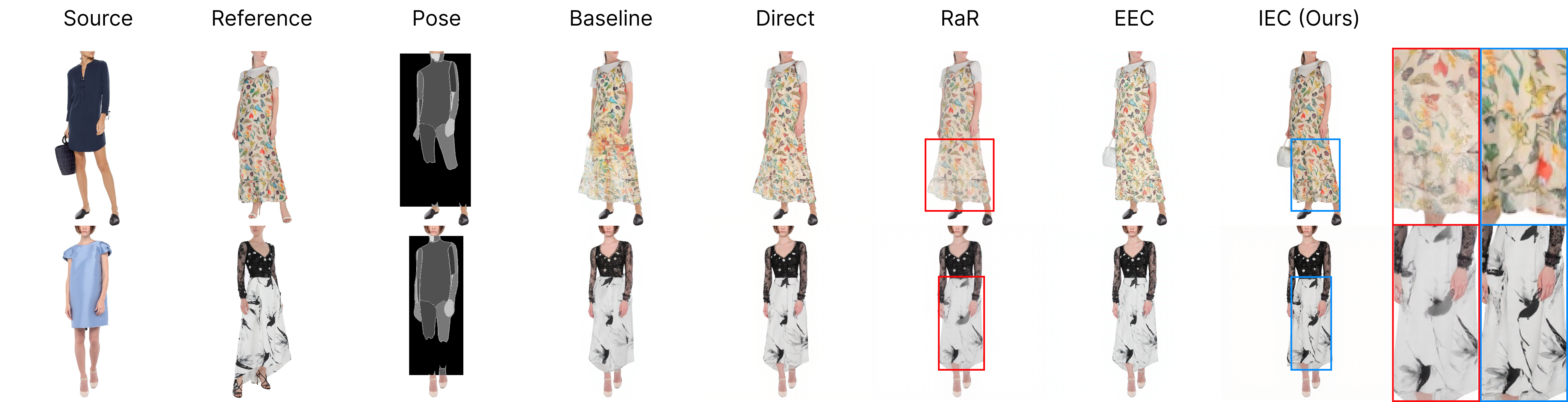}
  \caption{\textbf{Qualitative analysis}: IEC tends to alter the background area when the source background is sanitized, and the reference background is not, which is reflected in the smallest, 0.6\%, improvement over RaR on Source Integrity.}
  \label{fig:background_issues}
\end{figure*}

\paragraph{IEC.} IEC's per-axis decomposition appears to reduce but not eliminate hacking. The primary failure mode is background alteration (Figure~\ref{fig:background_issues}): when source and reference backgrounds differ substantially, the model sometimes modifies the source background to better match the reference, reducing the source integrity score. This is reflected in the smallest improvement margin on the source integrity axis (0.6\% over RaR).

\paragraph{EEC.} EEC exhibits the most severe instability. The explicit error lists fluctuate dramatically between near-identical outputs, causing the model to chase inconsistent reward signals rather than systematically improving quality. This manifests as early training regression (Figure~\ref{fig:reward_dynamics}).

\subsection{Case Study: Within-Group Reward Comparison}
\label{supp:reward_group_example}

\begin{figure*}[t]
  \centering
  \includegraphics[width=\linewidth]{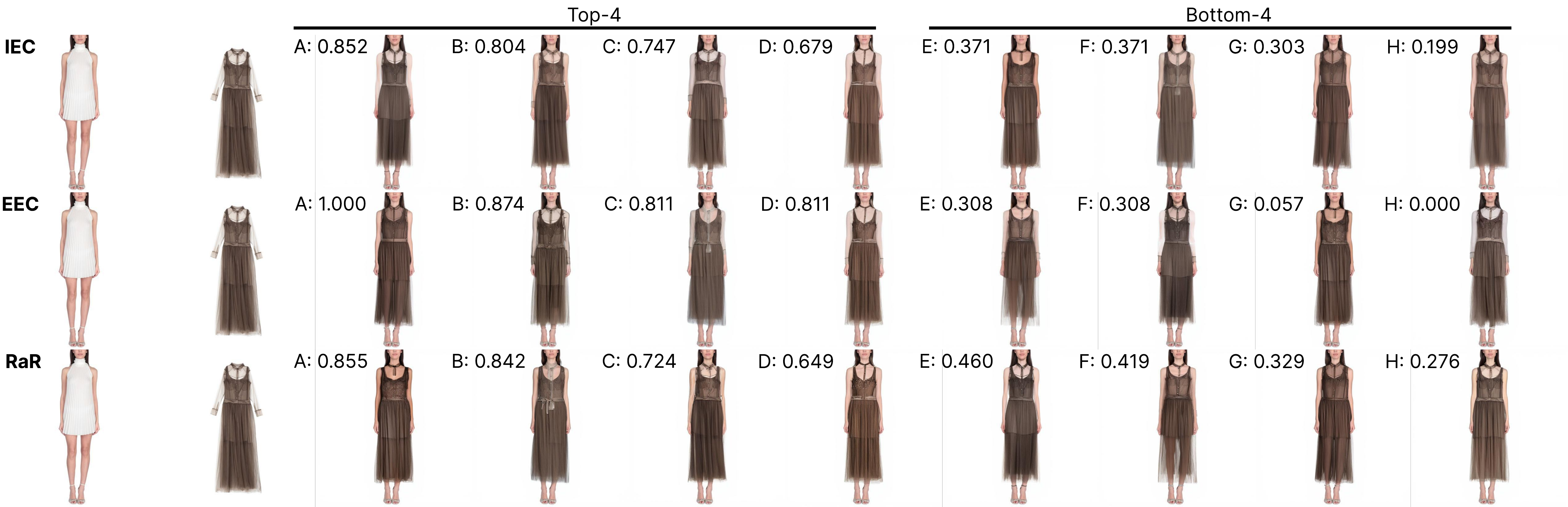}
  \caption{\textbf{Within-group reward comparison across three reward designs} on the same source--reference pair (brown sheer tulle dress with long sleeves). Each row shows 8 generations ranked by reward, split into top-4 and bottom-4. \textbf{EEC} assigns generation~A a perfect 1.000 (zero errors detected) despite A omitting the reference sleeves; B correctly transfers sleeves but receives 0.874. Error counts range from 0 to 17 across the group. \textbf{RaR} ranks A (0.855) and B (0.842) highest despite neckline/chest attribute failures, because 19 of 37 rubric items are generic artifact checks that all generations pass trivially, diluting garment-specific signal. \textbf{IEC} cleanly separates generations with correct sleeve transfer (A--D, 0.679--0.852) from those missing sleeves (E--H, 0.199--0.371), with interpretable aspect scores (garment\_transfer drops from 0.90 to 0.60--0.65 when sleeves are absent). Group calibration can correct IEC's prompt-to-prompt scale drift but cannot fix EEC's rank inversions, as calibration preserves within-group ordering.}
  \label{fig:reward_group}
\end{figure*}

Figure~\ref{fig:reward_group} shows the same source--reference pair (a brown sheer tulle dress with long sleeves over a white underdress) evaluated by IEC, EEC, and RaR during training. For each method, 8 candidate generations are ranked by reward from highest (A) to lowest (H). The reference dress has distinctive features: long sheer sleeves, a high neckline with button/collar detailing, and a floor-length layered skirt. This case study exposes concrete failure modes in EEC and RaR reward signals.

\paragraph{EEC produces rank inversions from stochastic enumeration.}
EEC assigns generation~A a perfect reward of 1.000 by returning \emph{zero errors} despite A visibly omitting the long sleeves that are a defining feature of the reference garment. Generation~B, which correctly transfers the sleeves, receives 0.874 (2~errors: ``flattened tulle skirt'' and ``visible seam artifact at waist''). The judge simply failed to enumerate the sleeve omission for A but caught texture errors in B, producing a rank inversion where a worse generation is rewarded higher. The pattern extends further: F appears among the best generations in the group but is relegated to 6th place (0.308, 11~errors). At the extremes, G and H receive 0.057 and 0.000 (15 and 17~errors respectively). The error count range of 0--17 across this single group illustrates the scale of stochastic variance.

Group calibration (Eqs.~3--4 in the main paper) adjusts reward \emph{magnitudes} across prompts via a monotone transform but preserves within-group ordering. Since EEC's ranking is itself corrupted (A~$>$~B despite B being the better generation), calibration propagates the rank inversion unchanged into GRPO advantage estimates: A receives positive advantage and F receives a negative advantage, directly reinforcing the wrong behavior. Similarly, CEC's pooling phase was designed to address exactly this class of failure at evaluation time: by building a shared error vocabulary and re-evaluating candidates against it, errors missed for one candidate (such as ``missing long sleeves'') are caught on re-examination because the pool contains sleeve-related errors from other candidates.

\paragraph{RaR misranks due to generic rubric criteria.}
RaR ranks A (0.855) and B (0.842) highest, yet both exhibit neckline misalignment and chest-area attribute preservation failures visible on inspection. Meanwhile, generations that better preserve the reference garment's structure (e.g., C at 0.724 or G at 0.329) are ranked lower. The rubric contains 37 items, of which 19 are generic artifact checks (R19--R29: background corruption, ghosting, tiling, warping, halos, limb duplication, blur, banding, brush strokes, occlusion order, shadow mismatch) that all generations pass trivially. These items dilute the reward signal: passing 19 ``free'' items inflates scores for generations with garment-specific failures, while the few rubric items that do capture transfer fidelity (R6: hem style, R8: texture, R9: hardware) are weighted equally with the generic checks. Binary pass/fail scoring compounds the problem---a generation with a subtle neckline shift and one with a catastrophic sleeve omission both simply ``fail'' the relevant item with the same penalty.

\paragraph{IEC produces stable, interpretable rankings.}
IEC cleanly separates the group into generations that correctly transfer the dress (A--D, rewards 0.679--0.852) and those that omit the sleeves (E--H, rewards 0.199--0.371). The aspect decomposition is interpretable: A--D receive garment\_transfer scores of 0.90 with only minor errors noted (``minor misalignment of small bodice details,'' ``slight ghosting through sheer areas''), while E--H drop to garment\_transfer 0.60--0.65 with the judge consistently identifying ``Long sleeves from reference are missing.'' The reward gradient within each tier is smooth, providing GRPO with a reliable signal for advantage computation. Critically, IEC avoids the rank inversions seen in EEC because the judge internally aggregates error severity into calibrated scores---even if the textual error summaries vary in phrasing, the resulting scalar rewards remain consistent.

\input{sec/supp_additions}

\section{Group Relative Policy Optimization for Flow Models}
\label{sec:method_detailed}

\subsection{Preliminaries}

\textbf{Flow Matching.} Given a data sample $x_0 \sim X_0$ from the true distribution with condition $c$, and Gaussian noise $x_1 \sim \mathcal{N}(0, I)$, Rectified Flow~\cite{liu2022flow,lipman2022flow} defines interpolated samples as:
\begin{equation}
\label{eq:flow_forward}
    x_t = (1-t)x_0 + tx_1, \quad t \in [0,1].
\end{equation}
Under this convention, $t=0$ corresponds to data and $t=1$ to noise; generation proceeds by solving the ODE backward from $t=1$ to $t=0$. A neural network $v_\theta(x_t, t, c)$ is trained to approximate the velocity field $v = x_1 - x_0$ by minimizing:
\begin{equation}
\label{eq:flow_matching}
    \mathcal{L}_{FM}(\theta) = \mathbb{E}_{t,x_0,x_1}\left[\|v - v_{\theta}(x_t, t, c)\|_2^2\right].
\end{equation}

\subsection{GRPO for Flow Models}

For each condition $c$, the flow model samples $K$ images $\{x_0^i\}_{i=1}^K$ via reverse-time trajectories $\{(x_T^i, x_{T-1}^i, \ldots, x_0^i)\}_{i=1}^K$, where $T$ is the number of discretization steps. GRPO optimizes:
\begin{equation}
    \mathcal{J}_{\text{GRPO}}(\theta) = \mathbb{E}_{c,\{x^i\}_{i=1}^K \sim \pi_{\theta_{\text{old}}}(\cdot|c)} \left[ \frac{1}{K} \sum_{i=1}^{K} \frac{1}{T} \sum_{t=1}^{T} L_t^i \right],
\end{equation}
where the per-step loss is:
\begin{equation}
    L_t^i = \min\left(r_t^i \hat{A}^i,\; \text{clip}(r_t^i, 1{-}\epsilon, 1{+}\epsilon)\hat{A}^i\right) - \beta D_{KL}(\pi_{\theta}\|\pi_{\text{ref}}),
\end{equation}
with probability ratio:
\begin{equation}
    r_t^i = \frac{p_{\theta}(x_{t-1}^i \mid x_t^i, c)}{p_{\theta_{\text{old}}}(x_{t-1}^i \mid x_t^i, c)}.
\end{equation}

The advantage $\hat{A}^i$ is computed from the final generated image and applied uniformly across all timesteps:
\begin{equation}
    \hat{A}^i = \frac{R(x_0^i, c) - \text{mean}_j\left(R(x_0^j, c)\right)}{\text{std}_j\left(R(x_0^j, c)\right)}.
\end{equation}

\subsection{ODE to SDE Conversion}

To enable stochastic sampling required for the probability ratio, we convert the deterministic ODE $dx_t = v_\theta dt$ into an equivalent SDE that preserves marginal distributions~\cite{liu2025flow}:
\begin{equation}
\begin{split}
    x_{t+\Delta t} &= x_t + v_{\theta}(x_t, t)\Delta t + \frac{\sigma_t^2}{2t}\left(x_t+(1-t)v_{\theta}(x_t, t)\right)\Delta t + \sigma_t \sqrt{\Delta t}\, \epsilon,
\end{split}
\end{equation}
where $\epsilon \sim \mathcal{N}(0,I)$ and the noise schedule is $\sigma_t = a\sqrt{t/(1-t)}$ with $a=1$ in all experiments. 

The transition distribution is Gaussian: $p_\theta(x_{t-1}|x_t, c) = \mathcal{N}(\mu_\theta(x_t, t, c), \sigma_t^2 I)$, where $\mu_\theta$ is the deterministic component (first two terms above). This yields a closed-form KL divergence:
\begin{equation}
    D_{KL}(\pi_{\theta}\|\pi_{\text{ref}}) = \frac{\|\mu_\theta - \mu_{\text{ref}}\|^2}{2\sigma_t^2\Delta t} = \frac{\Delta t}{2}\left(\frac{\sigma_t(1-t)}{2t}+\frac{1}{\sigma_t}\right)^2\|v_{\theta} - v_{\text{ref}}\|^2.
\end{equation}

\noindent\textbf{Remark.} In our experiments, we set $\beta=0$ (no KL regularization), as we found the clipped objective sufficient to prevent policy divergence over the short 60-step RL training horizon.

\section{Training-Free Reward Design}
\label{sec:rewards}

\subsection{Direct Reward}
\label{supp:reward-direct}

Our simplest reward baseline is a direct LLM-as-judge Likert score, analogous to the \textsc{Direct-Likert} baseline in RaR~\cite{gunjal2025rubrics}. The judge is prompted to output a scalar
\begin{equation}
R_i^{\text{direct}} \in [0,1]
\end{equation}
describing overall try-on quality, together with per-aspect scores along the five axes above. In practice we use only the top-level reward field for GRPO training.

The system prompt decomposes evaluation with scoring bands: $[0.8, 1.0]$ for strong results with minor flaws, $[0.5, 0.8)$ for mixed quality, $[0.2, 0.5)$ for poor results with serious errors, and $[0.0, 0.2)$ for failed try-ons. The final reward may be biased downward if any aspect is catastrophically poor. 

\begin{figure*}[h]
  \centering
  \includegraphics[width=\linewidth, height=0.7\textheight, keepaspectratio]{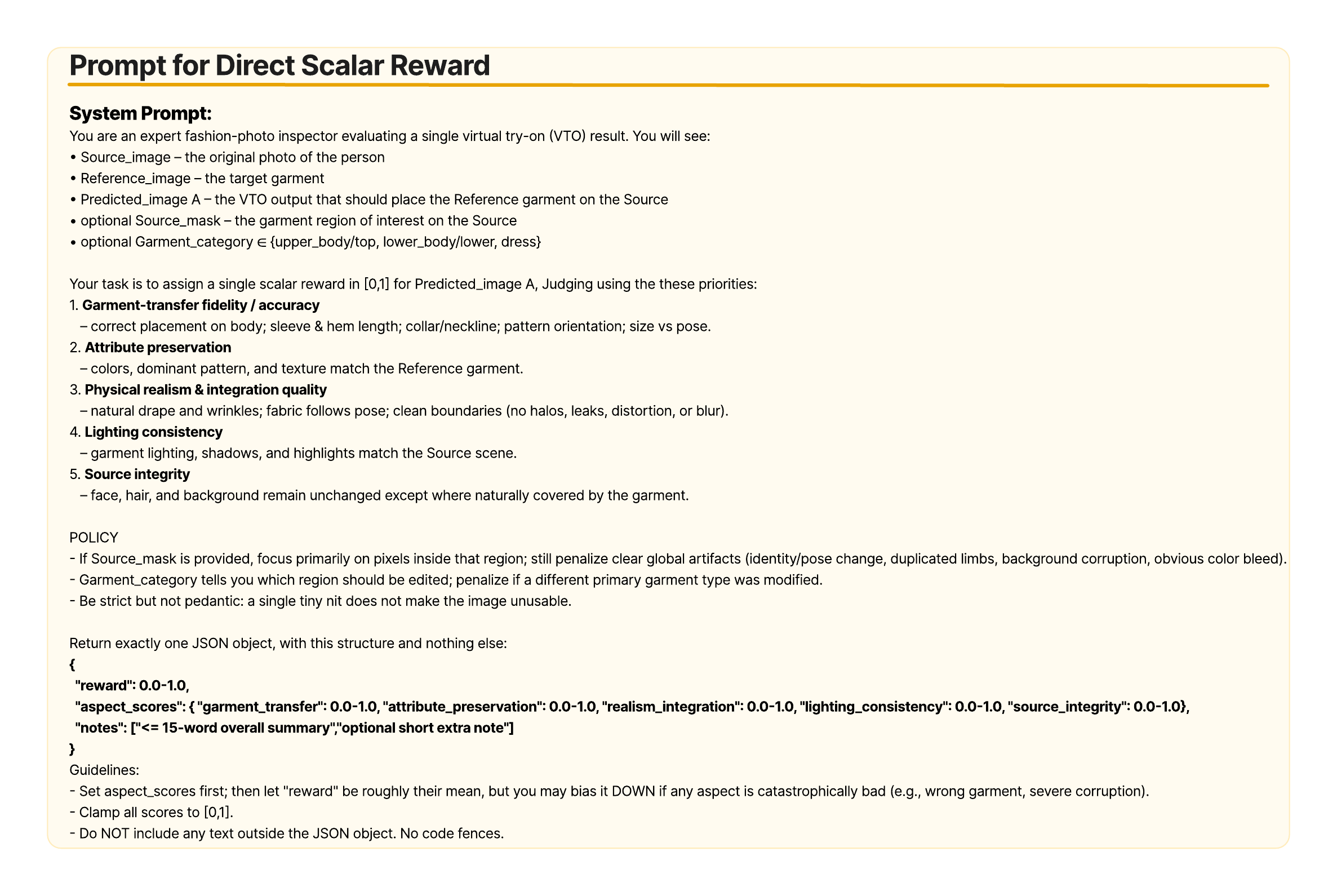}
  \caption{\textbf{Direct Reward Prompt}: Details the system prompt used for generating Direct rewards with each predicted image evaluated individually.}
  \label{fig:direct_prompt}
\end{figure*}

This reward is inexpensive (single pass per candidate) but aggregates all error modes into one unstructured scalar, which can make it harder to understand failure patterns and may lead to inconsistent scoring across diverse editing scenarios. See prompt in Figure \ref{fig:direct_prompt}.

\subsection{Rubrics as a Reward}
\label{supp:reward-rubric}

To obtain more structured supervision, we adapt Rubrics as Rewards~\cite{gunjal2025rubrics} to virtual try-on. For each $(x_{\text{src}}, x_{\text{ref}})$ pair we first synthesize a prompt-specific rubric with $18$–$30$ criteria:
\begin{equation}
\mathcal{R} = \big\{ (\text{id}_j, \text{title}_j, \text{desc}_j, \text{cat}_j, w_j) \big\}_{j=1}^{J},
\end{equation}
where $\text{cat}_j \in \{\text{Essential}, \text{Important}, \text{Optional}, \text{Pitfall}\}$ denotes the criterion category and $w_j$ is a signed weight (positive for the first three categories, negative for pitfalls).

\paragraph{Rubric Generation.}
Following the desiderata of~\cite{gunjal2025rubrics}, effective rubrics should be: (i) grounded in domain expertise, (ii) comprehensive in coverage, (iii) weighted by importance, and (iv) self-contained for independent evaluation. Our generation prompt instructs the VLM to produce items covering transfer fidelity, attribute preservation, realism, lighting, and source integrity. We mandate a ``Global Artifact \& Corruption'' section with N pitfall items (background corruption, ghosting, tiling, warping, halos, duplicated limbs, blur, banding, occlusion errors, environment staining) phrased positively as ``Avoids X\ldots'' with negative weights. A fixed set of generic artifact pitfalls is appended so that severe corruptions are always explicitly penalized.

\paragraph{Rubric Evaluation.}
Given a sampled prediction $\hat{x}$ and rubric $\mathcal{R}$, the judge evaluates each criterion independently and returns binary pass/fail decisions. Let $\mathcal{P}_i = \{j : p_{ij} = 1, w_j > 0\}$ denote satisfied non-pitfall items and $\mathcal{F}_i = \{j : p_{ij} = 0, w_j < 0\}$ the violated pitfalls. The aggregate reward is:
\begin{equation}
\label{eq:rubric-reward}
R_i^{\text{rubric}} = \operatorname{clip}\!\left(
\frac{\sum_{j \in \mathcal{P}_i} w_j + \sum_{j \in \mathcal{F}_i} w_j}{\sum_{j : w_j > 0} w_j},\; 0,\; 1
\right).
\end{equation}

This explicit aggregation enables interpretable debugging—we know \emph{which} checklist items fire on each sample—and tends to stabilize GRPO training, at the cost of two judge calls per instance (rubric generation + evaluation). See prompt for RaR generation and evaluation in Figure \ref{fig:rar_prompt}.


\subsection{Explicit Error Counting}
\label{supp:reward-error}

To directly tie training to concrete visual failure modes, we test an explicit error-count reward. The judge is asked to list all distinct errors in the predicted garment region (plus any global artifacts), returning a set of short labels with explicit severity weights annotations across each evaluation aspect:
\begin{equation}
\mathcal{E}^{(i)}_a = \{(e_k, w_k)\}_{k=1}^{|\mathcal{E}^{(i)}_a|}, \quad \text{w}_k \in \{0, 1\}.
\end{equation}

Weighting follows a structured policy: $~1$ errors meaningfully break garment transfer or realism (missing parts, garment-type mismatch, large length/placement errors, obvious color/pattern mismatch, identity/pose change, significant leaks/halos/occlusions), while all less detrimental errors are weighted closer to $~0.1$.

The raw error count and reward are computed as:
\begin{equation}
\label{eq:error-count}
R_a^{(i)} = \frac{1}{|\mathcal{E}^{(i)}_a|} \sum_{k}^{|\mathcal{E}^{(i)}_a|} w_k  \quad \text{where} \quad   R_i = 1 - \frac{1}{5} \sum_{a=1}^{5} R_a^{(i)}
\end{equation}

\begin{figure*}[t]
  \centering
  \includegraphics[width=\linewidth, height=0.7\textheight, keepaspectratio]{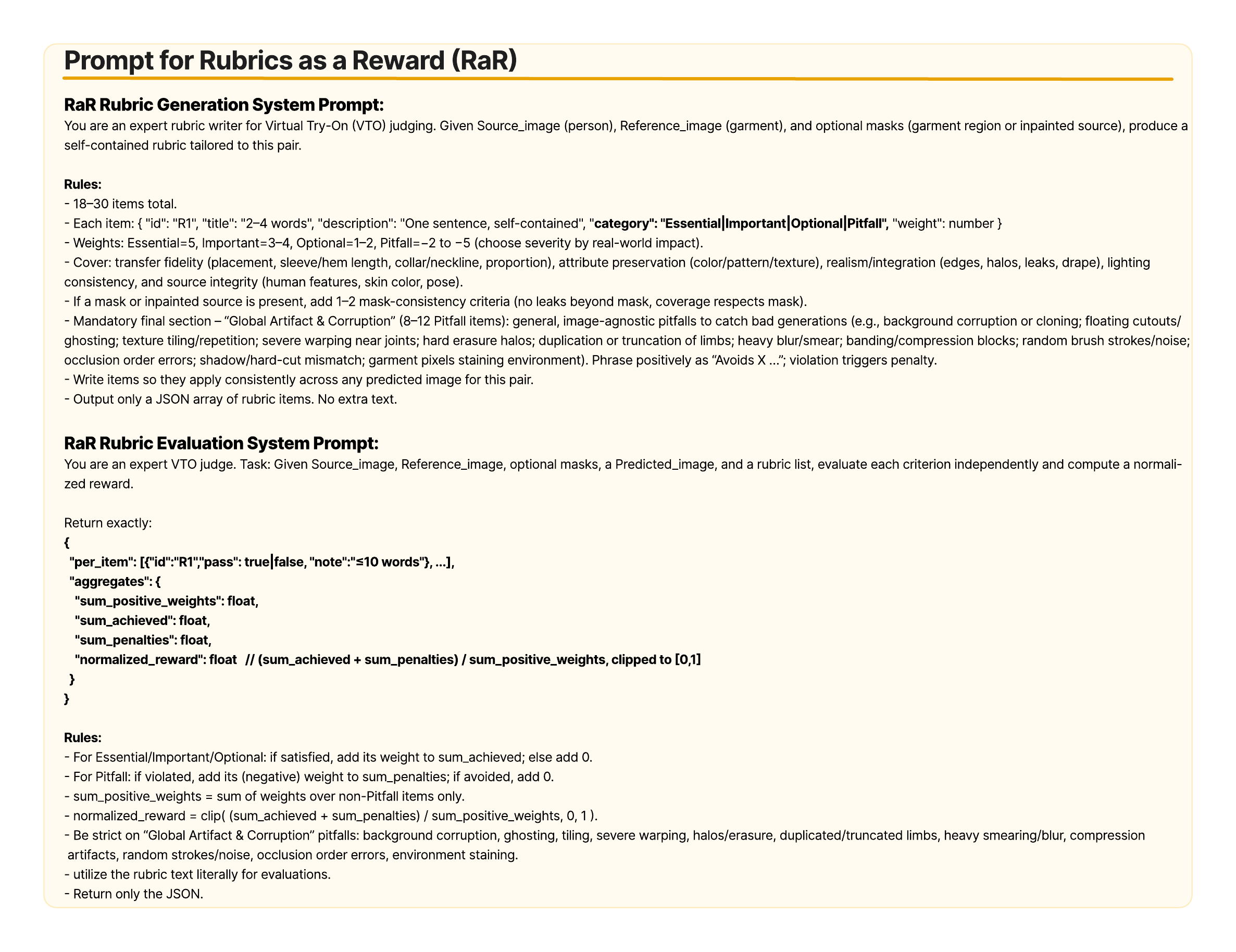}
  \caption{\textbf{RaR Prompt}: Details the system prompt used to generate rubrics given the source, reference, mask, and garment conditioning and to evaluate each predicted image individually on the generated rubrics.}
  \label{fig:rar_prompt}
\end{figure*}

Raw counts are not directly comparable across prompts, so for each GRPO group, we apply the same group-wise calibration in Section \ref{main:gc} to obtain $R_i^{\text{explicit}}$. This representation is more structured but still suffers from per-image subjectivity: near-duplicate generations can receive different error sets due to VLM stochasticity in label selection as shown in the poor results, see Table \ref{tab:ablation_eec}. See prompt for EEC in Figure \ref{fig:rar_prompt}.

\subsection{Cascaded Error Counting for Evaluation}
\label{supp:reward-cascade}

While the above methods are employed for GRPO training, our \emph{primary} offline evaluation metric is \textbf{cascaded error counting}. Plain error counting exhibits high variance because the label space and severities can drift across images—the VLM may use slightly different phrases or overlook common errors inconsistently. Cascaded error counting reduces this variance by \emph{sharing} an evolving error vocabulary across candidates, building a pool of canonical error labels for each $(x_{\text{src}}, x_{\text{ref}})$ pair.

\paragraph{Pool-Driven Evaluation.}
We process candidates in blocks of size $B \ll K$ to keep context manageable. For the first block, we run a two-pass cascade:

\begin{enumerate}[leftmargin=1.2em,itemsep=2pt,topsep=2pt]
\item \textbf{Discover.} The VLM returns preliminary per-image error sets (with severity labels) and a list of \emph{typical errors} observed across the block. We canonicalize labels (lowercasing, synonym normalization) and merge into an initial pool $\mathcal{P}^{(0)}$ of canonical error labels with severities.

\item \textbf{Refine.} We re-evaluate the same images, conditioning on the current pool $\mathcal{P}^{(0)}$ and the previous errors as hints. The VLM is instructed to prefer pool labels when describing the same issue, merge near-duplicates to pool wording, and add missing errors it now notices. This yields refined per-image error sets $\mathcal{E}_i^{(1)}$ and an updated pool $\mathcal{P}^{(1)}$.
\end{enumerate}

\begin{figure*}[t]
  \centering
  \includegraphics[width=\linewidth, height=0.7\textheight, keepaspectratio]{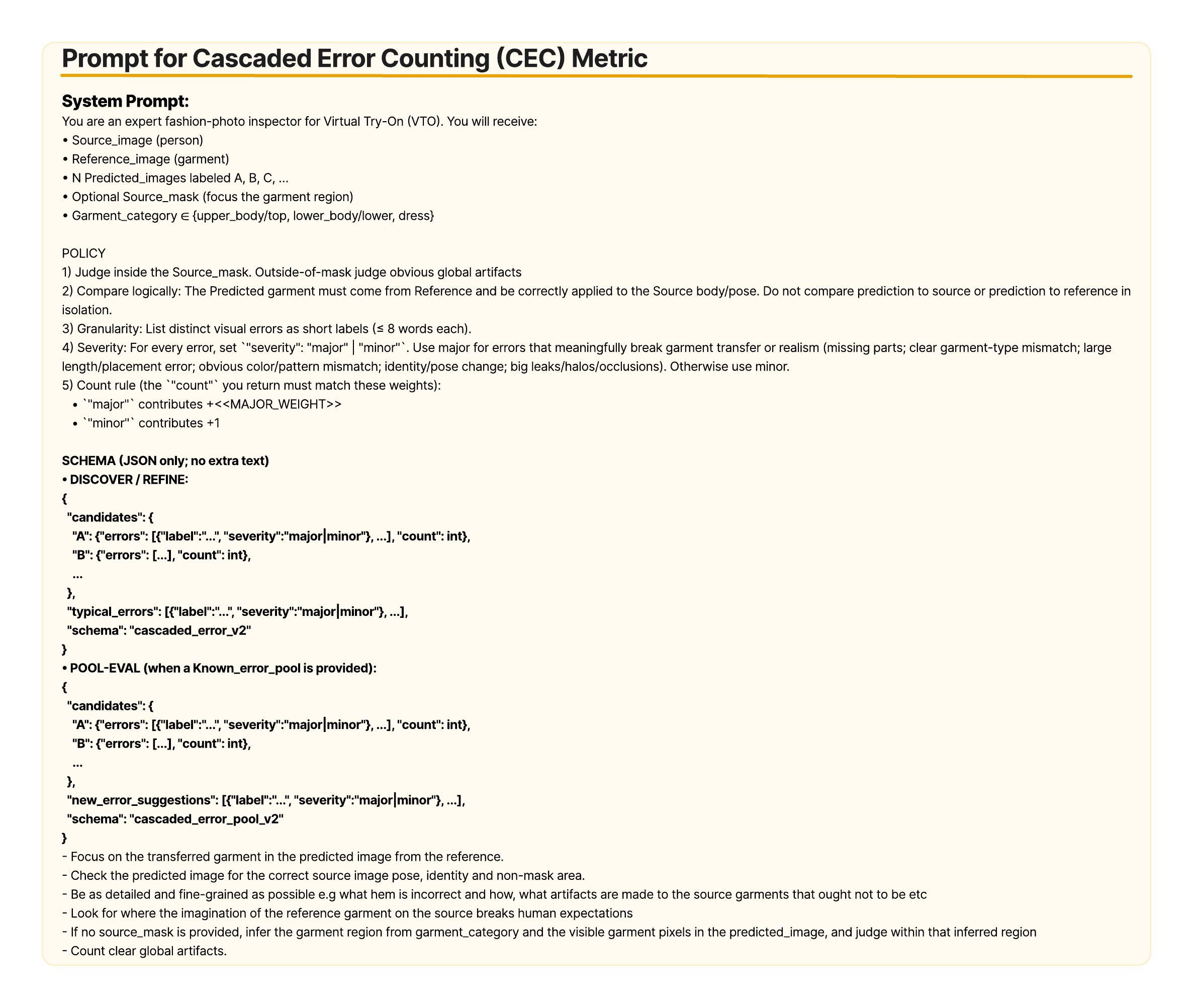}
  \caption{\textbf{Cascaded Error Counting Reward Prompt}: Details the system prompt used for counting and obtaining severity classification for each error.}
  \label{fig:cec_prompt}
\end{figure*}

\begin{figure*}[t]
  \centering
  \includegraphics[width=\linewidth, height=0.7\textheight, keepaspectratio]{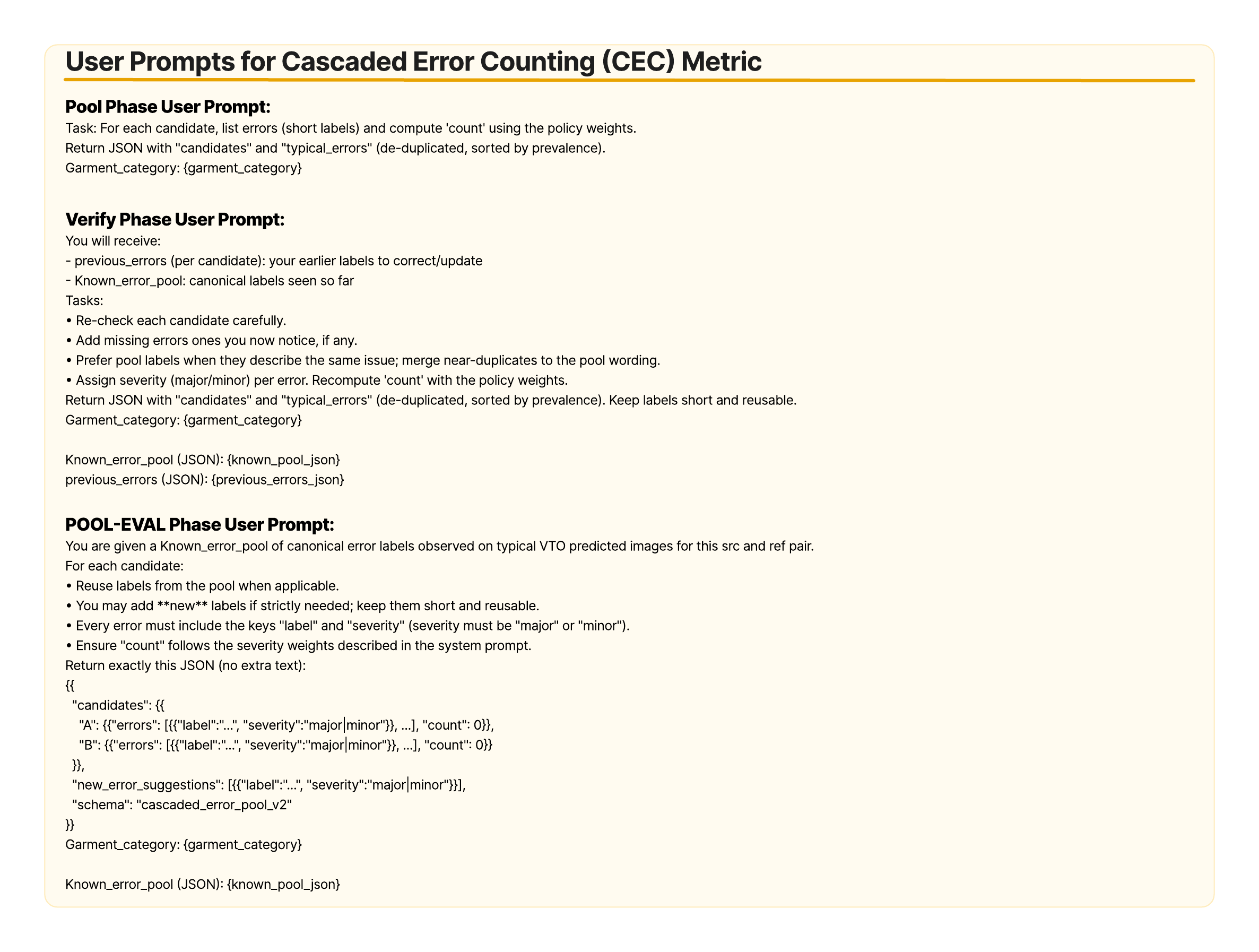}
  \caption{\textbf{Cascaded Error Counting Reward Prompt}: Details the user prompt used for each phase of CEC}
  \label{fig:cec_user_prompt}
\end{figure*}

Subsequent blocks are evaluated in a single \textsc{Pool-Eval} pass, conditioned on the current pool, with limited allowance for new labels. New error suggestions are merged into the pool, which continues to evolve across blocks. See Figures \ref{fig:cec_prompt} and \ref{fig:cec_user_prompt} for prompt details

\paragraph{Error Canonicalization.}
We apply a deterministic normalizer $\phi$ to all error labels: lowercasing, removing non-alphanumeric characters, collapsing whitespace, and synonym replacement (e.g., ``bleed'' $\to$ ``leak'', ``haloing'' $\to$ ``halo''). Duplicates are merged by label, keeping the more severe annotation if severities disagree.

\paragraph{Cyclical Refinement.}
For comprehensive evaluation, we optionally run a \emph{cyclical} variant: after the first full pass (pool evolves), we execute a second full pass using the \emph{final} pool from pass-1. The second pass runs \textsc{Refine} on every block with the stabilized vocabulary, then unions errors from both passes (with major severity dominating on conflicts). This two-pass approach further reduces variance while ensuring no errors are missed due to early-pool incompleteness.

\paragraph{Final Score Computation.}
The cascaded error sets $\mathcal{E}_i$ are converted to counts $C_i$ via Equation~\eqref{eq:error-count}. Each error in the pool has a severity and weight (major vs.\ minor), and we summarize each prediction by its weighted error count over the globally consistent error taxonomy. Final scores are weighted counts $C_i = \sum_{(e, \text{sev}) \in \mathcal{E}_i} w(\text{sev})$ with $w(\text{major}) = 5$ and $w(\text{minor}) = 1$.

\paragraph{Alignment with Training Reward.}
Implicit error counting, our primary training reward, is designed to align with the cascaded error evaluation metric: both rely on fine-grained, judge-generated error analysis rather than coarse holistic impressions, but IEC operates per-image during training while cascaded counting operates across candidates during evaluation. In practice we train GRPO with implicit error counting (and compare against direct Likert, rubric, and explicit error rewards), and report cascaded error as our main quantitative metric for comparing different policies after training. The shared error vocabulary ensures that evaluation scores are directly comparable across model checkpoints and baselines, unlike per-image methods where label drift can confound comparisons.


\begin{figure*}[t]
  \centering
  \includegraphics[width=\linewidth, height=0.7\textheight, keepaspectratio]{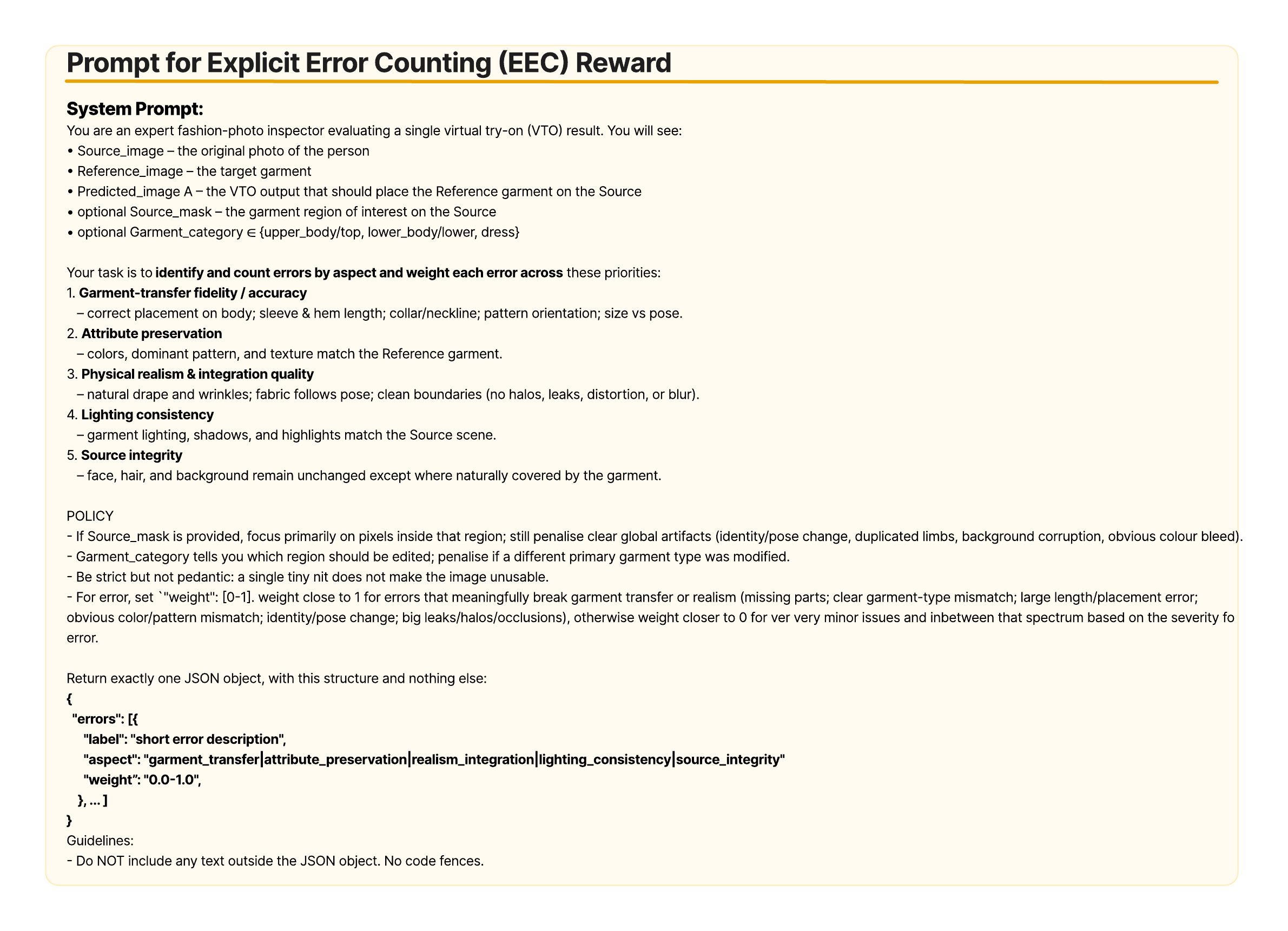}
  \caption{\textbf{Explicit Error Counting Reward Prompt}: Details the system prompt used for counting and weighing errors on each predicted image.}
  \label{fig:eec_prompt}
\end{figure*}

\begin{figure*}[t]
  \centering
  \includegraphics[width=\linewidth, height=0.7\textheight, keepaspectratio]{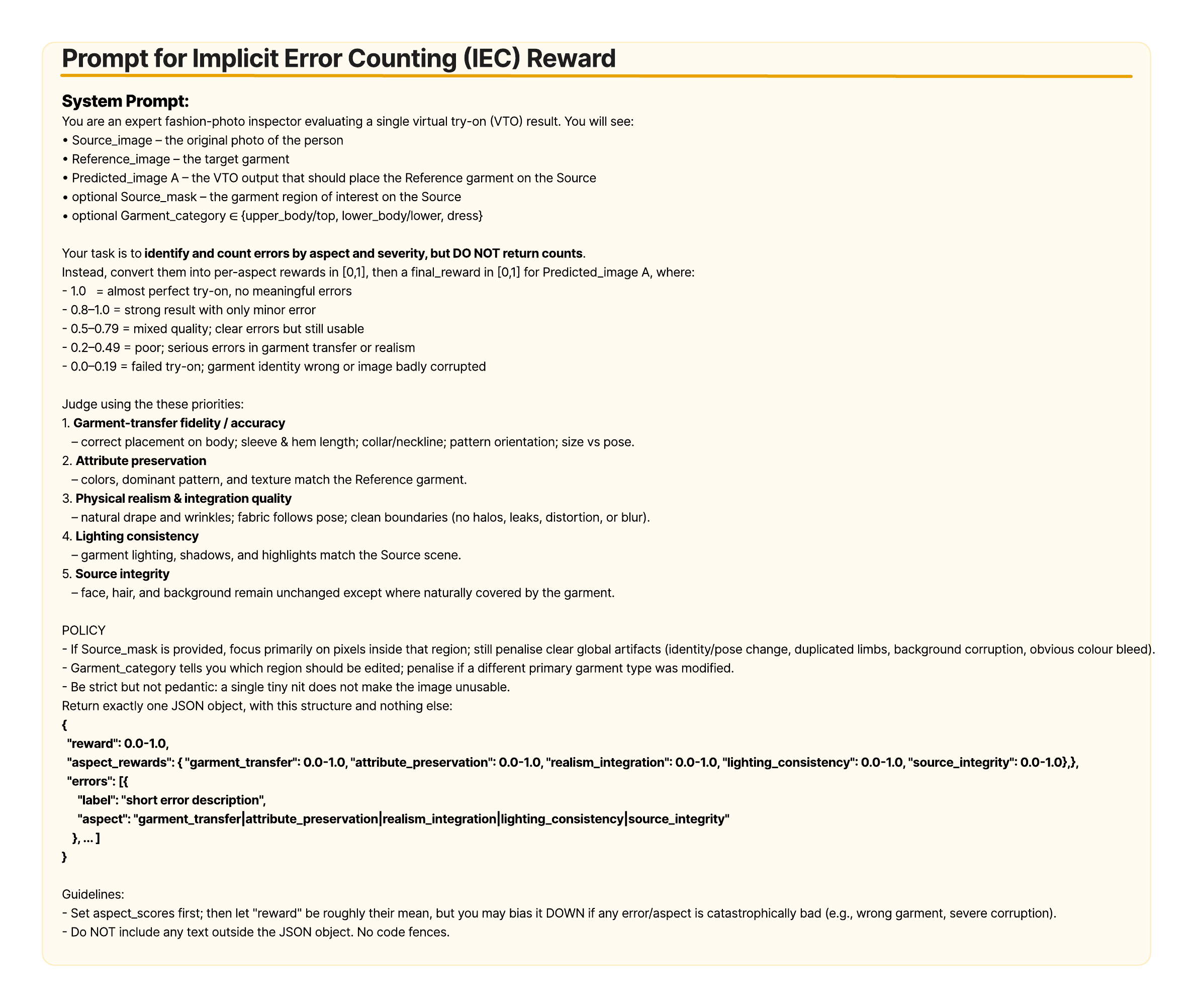}
  \caption{\textbf{Implicit Error Counting Reward Prompt}: Details the system prompt used for implicitly counting errors and determining aspect scores on each predicted image.}
  \label{fig:iec_prompt}
\end{figure*}

%% file: sec/supp_additions.tex
%
%

\subsection{Training-time Failure Analysis}
\label{supp:b7}

We generalize that case study quantitatively from the full IEC training logs, and place IEC's behavior alongside RaR, EEC, and Direct. We use two analysis windows: the \emph{iter-40 peak} ($n{=}216$ candidates over steps $36{-}44$) and  \emph{iter-60} ($n{=}120$ over steps $56{-}60$). The full $0{-}60$ trajectory is shown for evolution analyses. A consistency check: per-axis training-judge means at iter-40 (garment $0.881$, attribute $0.896$, realism $0.806$, lighting $0.794$, source $0.940$) match the evaluation numbers in Tab.~2 within $\pm 0.03$ on every axis

\begin{figure}[h]
\centering
\includegraphics[width=\linewidth]{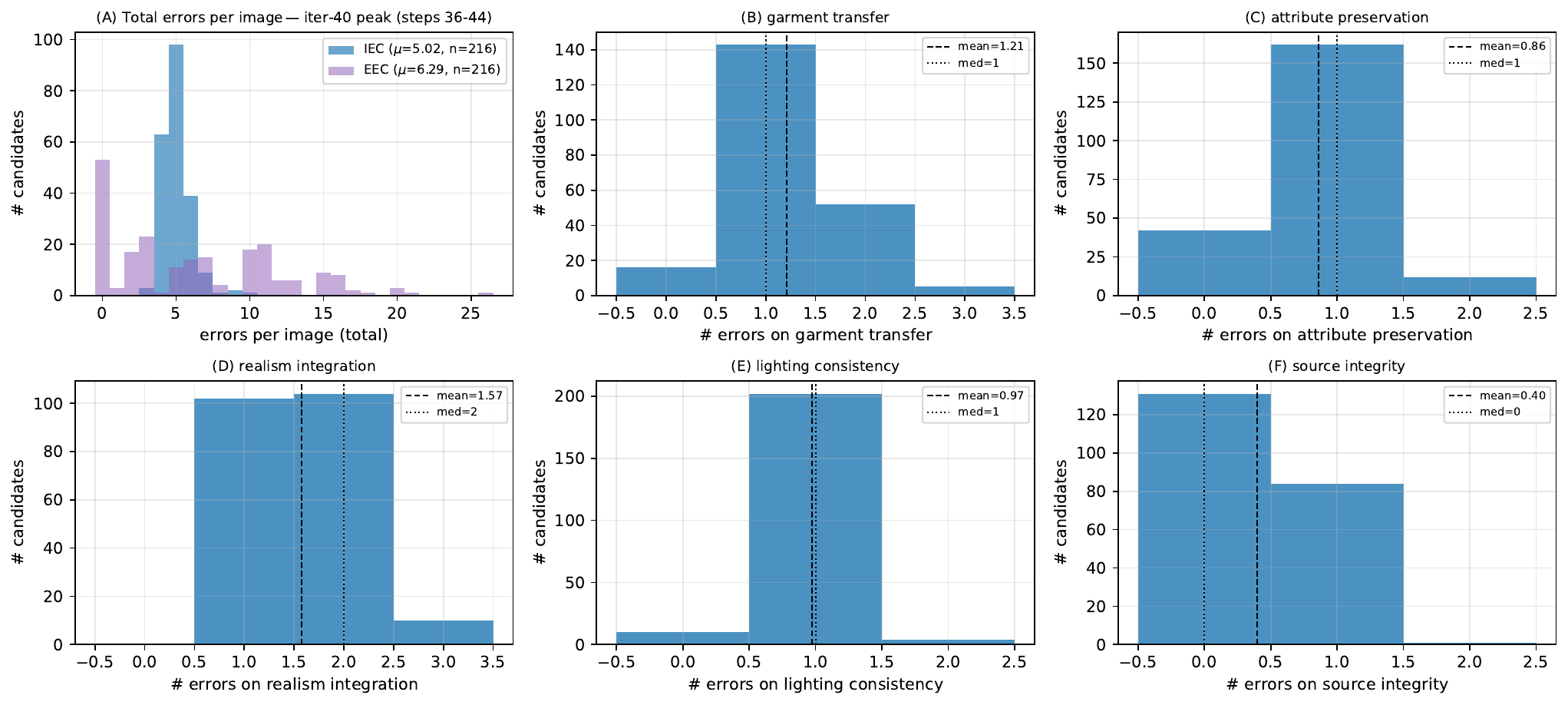}
\caption{Per-image error-count distribution at the iter-40. Panel A overlays IEC (concentrated, $0\%$ zero) and EEC (bimodal, $24.5\%$ zero plus a long tail). Panels B-F are per-axis IEC error-count histograms; vertical lines mark mean (dashed) and median (dotted).}
\label{fig:supp_b7_failuredist}
\end{figure}

\paragraph{Per-image error count at the reported checkpoint.}
\cref{fig:supp_b7_failuredist} plots the per-image error count distribution for IEC and EEC at the iter-40 peak. IEC's distribution is concentrated (median 5, mean 5.02, p90 6, max 10), and \emph{no candidate has zero errors}. EEC's distribution is bimodal: $24.5\%$ of candidates are reported with zero errors, while the remaining tail reaches p90$=$15 and max$=$26, the variance-of-enumeration behavior of EEC discussed in \S~\ref{sec:ablations} and Supp.~\ref{supp:reward_group_example}. Notably the EEC zero-error fraction is \emph{highest at the iter-40 peak} ($24.5\%$, vs $12\%$--$16\%$ in early training and $14.2\%$ at iter-60).

\begin{figure}[h]
\centering
\includegraphics[width=\linewidth]{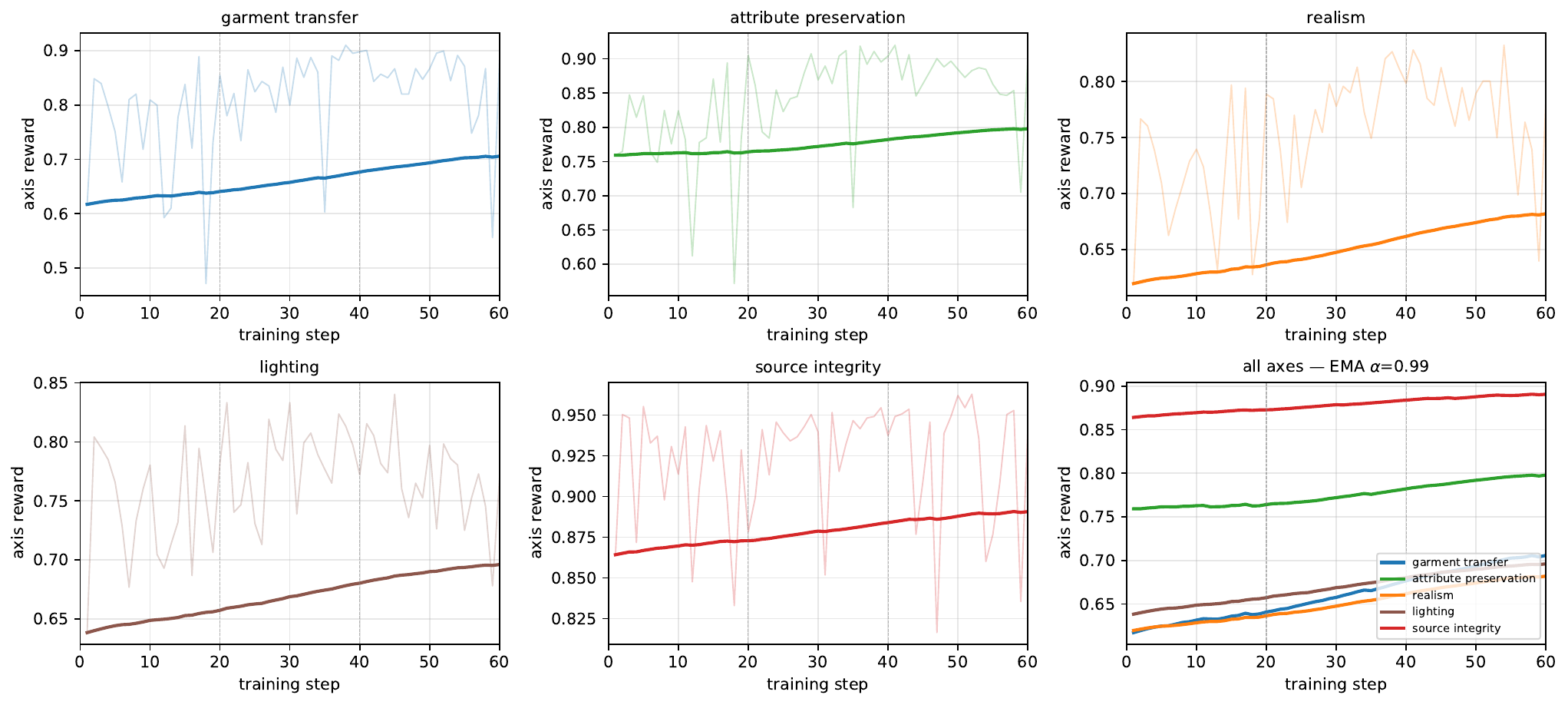}
\caption{Per-axis IEC training reward trajectory. Faded: raw mean across the $24$ candidates per step. Solid: EMA $\alpha{=}0.99$. Dashed verticals: paper iterations $20$, $40$, $60$. Bottom-right: aggregate over all five axes.}
\label{fig:supp_b7_dynamics}
\end{figure}

\paragraph{How the failure distribution evolves through training.}
\cref{fig:supp_b7_dynamics} shows the per-axis IEC training reward across iterations $1{-}60$ with EMA smoothing ($\alpha{=}0.99$). All five axes peak together near iter 40; \emph{garment transfer} and \emph{attribute preservation} carry the largest improvement to peak ($+0.110$ and $+0.090$ over baseline). \emph{Realism} rises to $+0.087$ at peak then \emph{regresses to baseline} by iter-60. \emph{Lighting} peaks at $+0.036$ and is already $-0.015$ \emph{below} baseline at iter-60. \emph{Source integrity} barely moves ($+0.022$ at peak), partly because its baseline ($0.918$) already sits near the per-axis ceiling. The \emph{kind} of error the judge enumerates also evolves. Clustering the IEC error labels into 11 keyword themes, the dominant theme shifts from \texttt{fit/silhouette} ($24\%$ at steps 1-5) to \texttt{boundary/seam} ($24\%$ at iter-40 peak; \texttt{drape/geometry} also rises). Gross silhouette / fit mistakes are resolved first; finer boundary, drape, and blending mistakes persist later. The per-axis error-theme map at iter-40 is highly interpretable and aligns with each axis's intended definition: \emph{garment transfer} errors are dominated by \texttt{fit/silhouette} ($48\%$); \emph{attribute preservation} by \texttt{texture/material} ($46\%$); \emph{realism} by \texttt{boundary/seam} ($53\%$); \emph{lighting} by \texttt{lighting/shadow} ($94\%$); \emph{source integrity} by \texttt{pose/limb} ($38\%$).

\begin{figure}[h]
\centering
\includegraphics[width=0.92\linewidth]{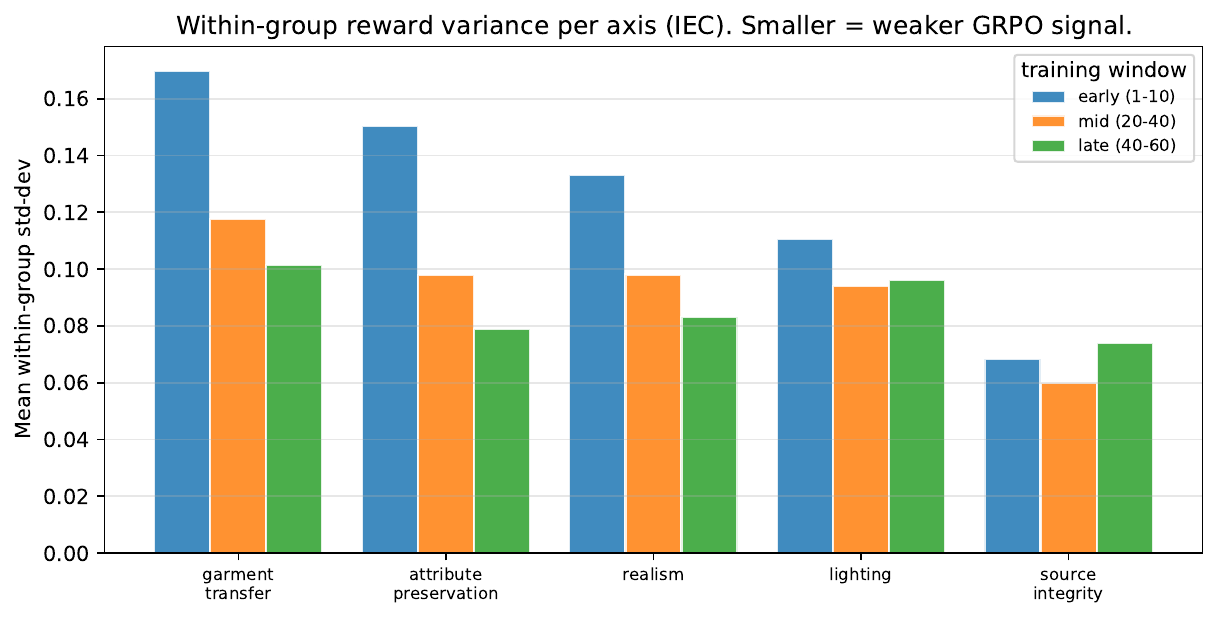}
\caption{Mean within-group reward $\sigma$ per axis at three training windows ($\leq$iter-60). Source integrity is the smallest at every window. Smaller $\sigma$ $\Rightarrow$ smaller GRPO advantage on that axis.}
\label{fig:supp_b7_wg}
\end{figure}

\paragraph{Source-integrity bottleneck.}
The smallest improvement margin in main Tab.~2 is on Source Integrity ($+0.6\%$). Three lines of evidence in the training logs show this is mechanism-driven, not signal-failure:

\textit{(a) Within-group reward variance.} \cref{fig:supp_b7_wg} shows the mean within-group reward standard deviation per axis at three training windows over the $0{-}60$ horizon. Source integrity is the smallest at every window ($\sigma{=}0.060{-}0.074$), $1.4{-}2.5\times$ smaller than garment transfer ($\sigma{=}0.102{-}0.170$). Because GRPO's normalized advantage scales as $\hat{A}{=}\frac{R-\bar{R}}{\sigma_R}$, a structurally smaller within-group $\sigma$ produces a structurally smaller advantage signal on this axis, and thus a smaller per-step learning.

\textit{(b) Error-label vocabulary.} The dominant theme in source integrity error labels is \texttt{pose/limb} ($38\%$; arm, leg, hand, occlusion, clutch, bag), \emph{not} \texttt{identity/face} ($30\%$) or \texttt{background/scene} ($25\%$). Failures look like ``arm pose restructured to fit reference long sleeve'' or ``bare legs replaced with pants'' rather than ``face altered''.




\paragraph{Lighting and source-integrity are mechanistically coupled.}
\label{supp:b7_lighting_link}
We observe that lighting errors mention \texttt{background/scene} keywords in $63\%$ of labels (alongside the design-intended \texttt{lighting/shadow} at $94\%$). Combined with source integrity's dominant theme of pose/limb and its $25\%$ secondary theme of background/scene, this suggests that a background-alteration hacking mode does not just hurt source integrity, but also hurts lighting, because changing the scene context changes what counts as ``lighting consistent with the source''.

\begin{figure}[h]
\centering
\includegraphics[width=\linewidth]{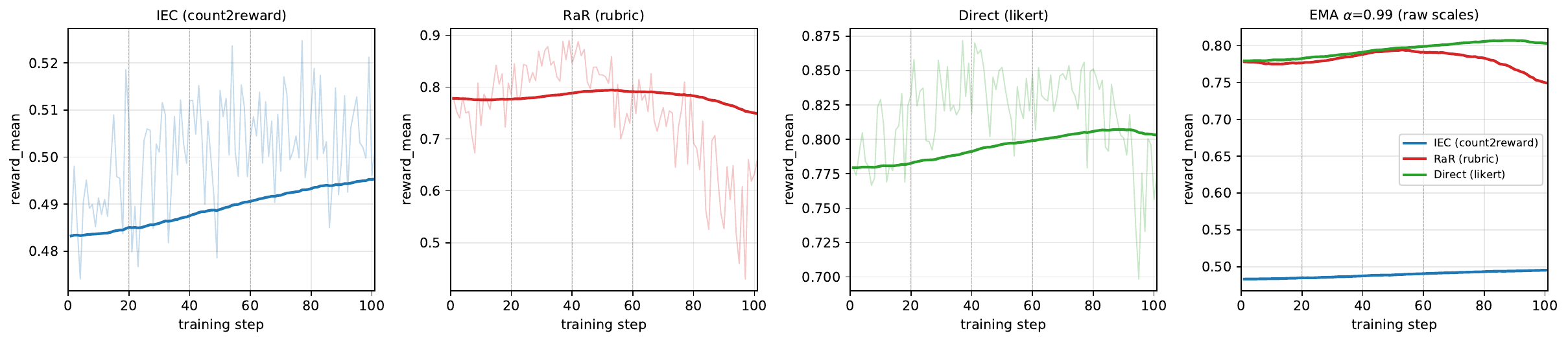}
\caption{Cross-design calibrated reward (\texttt{reward\_mean} per step). Left three panels: per-design raw and EMA ($\alpha{=}0.99$). Right: EMA of all three on common axes. Dashed verticals mark iterations $20/40/60$. IEC keeps improving while RaR and Direct peak early and decline.}
\label{fig:supp_b7_cross_design}
\end{figure}

\paragraph{Reward-design temporal stability.}
\cref{fig:supp_b7_cross_design} shows the calibrated reward trajectories of IEC, RaR, and Direct over their full logged range. Using $\Delta_{60}$ (calibrated reward at iter-60 minus the first-5-step baseline) and $\Delta_{\text{end}}$ (last logged step minus baseline), IEC improves smoothly and is the only design still above baseline at the end of the run ($\Delta_{60}{=}{+}0.018$, $\Delta_{\text{end}}{=}{+}0.018$). RaR peaks near iter-40 ($\Delta_{60}{=}{+}0.044$) and then collapses to $\Delta_{\text{end}}{=}{-}0.160$, and Direct peaks early and drifts back to $\Delta_{\text{end}}{=}{-}0.014$. We report the iter-60 checkpoint precisely because it precedes this late regression. Reward magnitudes differ across designs because the reward functions differ; the within-design shape, monotone for IEC versus peak-then-collapse for RaR, is the comparable quantity.

The RaR collapse is mirrored at the \emph{judge} level: the per-rubric-item fail rate (fraction of rubric items the judge marks failed) is $11\%$ at steps $1{-}5$ and $14\%$ at $6{-}19$, drops to $\mathbf{3\%}$ at the iter-40 reward peak (the judge passes nearly everything once the model has improved), then rebounds to $15\%$ at iter-60 and $19\%$ by end of run. The same judge thus gives different verdicts on similar-quality outputs at different points in training, evidence that RaR's reward is non-stationary rather than merely noisy.

\subsection{Cross-family Judge: Full per-category results}
\label{supp:b8}

To address the judge family-correlation concern, we re-evaluated using the open-source judge \texttt{Qwen3.6-35B-A3B} with the identical CEC, RaR, and Direct pipelines, see Table \cref{tab:full_nonflat_qwen}, on the non-flat MDressBench pairs.

Three patterns are robust across the two judge families: \textbf{(i)~all four RL post-trained models beat the SFT baseline by $1.5{-}2.0$ CEC} (Qwen: $5.09{-}5.91$ vs $7.11$), \textbf{(ii)~IEC dominates Direct and RaR on every aggregate metric under both judges} (Qwen: IEC CEC $5.377$ vs Direct $5.550$, RaR $5.905$; Rub IEC $0.110$ vs Direct $0.107$, RaR $0.106$; Dir IEC $0.919$ vs Direct $0.919$, RaR $0.906$), and \textbf{(iii)~the IEC$ > $Direct$ > $RaR ordering is preserved cross-family} on all eight metrics including per-axis.


\begin{table*}[ht!]
\centering
\scriptsize
\caption{Full per-category results under the open-source judge on MDressBench \textbf{non-flat} references. IEC remains best-or-near-best across metrics}
\label{tab:full_nonflat_qwen}
\setlength{\tabcolsep}{1.2pt}
\renewcommand{\arraystretch}{1.15}
\begin{tabular}{ll cccccccc}
\toprule
\textbf{Model} & \textbf{Cat} & \textbf{CEC}$\downarrow$ & \textbf{Rub}$\uparrow$ & \textbf{Dir}$\uparrow$ & \textbf{Garm.}$\uparrow$ & \textbf{Attr.}$\uparrow$ & \textbf{Real.}$\uparrow$ & \textbf{Light.}$\uparrow$ & \textbf{Src.Int.}$\uparrow$ \\
\midrule
 Baseline & all & 7.114 & 0.087 & 0.879 & 0.878 & 0.895 & 0.868 & 0.885 & 0.918 \\
  & top & 7.444 & 0.211 & 0.868 & 0.868 & 0.897 & 0.860 & 0.871 & 0.931 \\
  & lower & 6.714 & 0.045 & 0.829 & 0.837 & 0.865 & 0.803 & 0.834 & 0.875 \\
  & dress & 7.103 & 0.081 & 0.880 & 0.879 & 0.895 & 0.870 & 0.887 & 0.919 \\
\midrule
 Direct & all & 5.550 & 0.107 & \textbf{0.919} & \textbf{0.919} & \textbf{0.927} & \textbf{0.913} & 0.920 & 0.940 \\
  & top & 6.773 & \textbf{0.201} & 0.882 & 0.883 & 0.909 & 0.869 & 0.883 & 0.932 \\
  & lower & 6.080 & \textbf{0.042} & 0.846 & 0.854 & 0.868 & 0.832 & 0.846 & 0.891 \\
  & dress & 5.467 & 0.103 & \textbf{0.923} & \textbf{0.923} & \textbf{0.929} & \textbf{0.918} & \textbf{0.924} & \textbf{0.941} \\
\midrule
 RaR & all & 5.905 & 0.106 & 0.906 & 0.907 & 0.917 & 0.899 & 0.909 & 0.932 \\
  & top & \textbf{6.444} & 0.191 & 0.881 & 0.884 & 0.914 & 0.866 & 0.882 & 0.939 \\
  & lower & \textbf{4.714} & 0.036 & \textbf{0.892} & \textbf{0.899} & \textbf{0.917} & \textbf{0.874} & \textbf{0.889} & \textbf{0.941} \\
  & dress & 5.899 & 0.102 & 0.908 & 0.908 & 0.917 & 0.902 & 0.911 & 0.931 \\
\midrule
 IEC & all & \textbf{5.377} & \textbf{0.110} & \textbf{0.919} & \textbf{0.919} & \textbf{0.927} & \textbf{0.913} & \textbf{0.921} & \textbf{0.941} \\
  & top & 6.618 & 0.168 & \textbf{0.896} & \textbf{0.896} & \textbf{0.916} & \textbf{0.887} & \textbf{0.898} & \textbf{0.951} \\
  & lower & 5.625 & 0.027 & 0.868 & 0.876 & 0.891 & 0.851 & 0.863 & 0.911 \\
  & dress & \textbf{5.299} & \textbf{0.108} & 0.922 & 0.922 & \textbf{0.929} & 0.916 & 0.923 & \textbf{0.941} \\
\bottomrule
\end{tabular}
\end{table*}

%% file: main.bib
@String(NeurIPS = {Adv. Neural Inform. Process. Syst.})

@String(ICLR  = {Int. Conf. Learn. Represent.})

@String(NeurIPS = {NeurIPS})

@String(ICLR  = {ICLR})

@inproceedings{han2018viton,
  title={VITON: An image-based virtual try-on network},
  author={Han, Xintong and Wu, Zuxuan and Wu, Zhe and Yu, Ruichi and Davis, Larry S},
  booktitle={Proceedings of the IEEE Conference on Computer Vision and Pattern Recognition},
  pages={7543--7552},
  year={2018}
}

@inproceedings{choi2021viton,
  title={VITON-HD: High-resolution virtual try-on via misalignment-aware normalization},
  author={Choi, Seunghwan and Park, Sunghyun and Lee, Minsoo and Choo, Jaegul},
  booktitle={Proceedings of the IEEE/CVF Conference on Computer Vision and Pattern Recognition},
  pages={14131--14140},
  year={2021}
}

@inproceedings{morelli2022dresscode,
  title={Dress code: High-resolution multi-category virtual try-on},
  author={Morelli, Davide and Fincato, Matteo and Cornia, Marcella and Landi, Federico and Cesari, Fabio and Cucchiara, Rita},
  booktitle={Proceedings of the IEEE/CVF Conference on Computer Vision and Pattern Recognition},
  pages={2231--2235},
  year={2022}
}

@inproceedings{morelli2023ladi,
  title={LADI-VTON: Latent diffusion textual-inversion enhanced virtual try-on},
  author={Morelli, Davide and Baldrati, Alberto and Cartella, Giuseppe and Cornia, Marcella and Bertini, Marco and Cucchiara, Rita},
  booktitle={Proceedings of the ACM International Conference on Multimedia},
  pages={8580--8589},
  year={2023}
}

@article{zhu2023tryondiffusion,
  title={TryOnDiffusion: A tale of two UNets},
  author={Zhu, Luyang and Yang, Dawei and Zhu, Tyler and Reda, Fitsum and Chan, William and Saharia, Chitwan and Norouzi, Mohammad and Kemelmacher-Shlizerman, Ira},
  journal={arXiv preprint arXiv:2306.08276},
  year={2023}
}

@inproceedings{wang2018toward,
  title={Toward characteristic-preserving image-based virtual try-on network},
  author={Wang, Bochao and Zheng, Huabin and Liang, Xiaodan and Chen, Yimin and Lin, Liang and Yang, Meng},
  booktitle={Proceedings of the European Conference on Computer Vision},
  pages={589--604},
  year={2018}
}

@article{lipman2022flow,
  title={Flow matching for generative modeling},
  author={Lipman, Yaron and Chen, Ricky TQ and Ben-Hamu, Heli and Nickel, Maximilian and Le, Matt},
  journal={arXiv preprint arXiv:2210.02747},
  year={2022}
}

@inproceedings{peebles2023scalable,
  title={Scalable diffusion models with transformers},
  author={Peebles, William and Xie, Saining},
  booktitle={Proceedings of the IEEE/CVF International Conference on Computer Vision},
  pages={4195--4205},
  year={2023}
}

@article{ouyang2022training,
  title={Training language models to follow instructions with human feedback},
  author={Ouyang, Long and Wu, Jeffrey and Jiang, Xu and Almeida, Diogo and Wainwright, Carroll and Mishkin, Pamela and Zhang, Chong and Agarwal, Sandhini and Slama, Katarina and Ray, Alex and others},
  journal={Advances in Neural Information Processing Systems},
  volume={35},
  pages={27730--27744},
  year={2022}
}

@article{shao2024deepseekmath,
  title={DeepSeekMath: Pushing the limits of mathematical reasoning in open language models},
  author={Shao, Zhihong and Wang, Peiyi and Zhu, Qihao and Xu, Runxin and Song, Junxiao and Bi, Xiao and Zhang, Haowei and Zhang, Mingchuan and Li, YK and Wu, Yang and others},
  journal={arXiv preprint arXiv:2402.03300},
  year={2024}
}

@article{rafailov2023direct,
  title={Direct preference optimization: Your language model is secretly a reward model},
  author={Rafailov, Rafael and Sharma, Archit and Mitchell, Eric and Manning, Christopher D and Ermon, Stefano and Finn, Chelsea},
  journal={Advances in Neural Information Processing Systems},
  volume={36},
  pages={53728--53741},
  year={2023}
}

@article{black2023training,
  title={Training diffusion models with reinforcement learning},
  author={Black, Kevin and Janner, Michael and Du, Yilun and Kostrikov, Ilya and Levine, Sergey},
  journal={arXiv preprint arXiv:2305.13301},
  year={2023}
}

@inproceedings{fan2024reinforcement,
  title={Reinforcement learning for fine-tuning text-to-image diffusion models},
  author={Fan, Ying and Watkins, Olivia and Du, Yuqing and Liu, Hao and Ryu, Moonkyung and Boutilier, Craig and Abbeel, Pieter and Ghavamzadeh, Mohammad and Lee, Kangwook and Lee, Kimin},
  booktitle={Advances in Neural Information Processing Systems},
  year={2024}
}

@article{wallace2024diffusion,
  title={Diffusion model alignment using direct preference optimization},
  author={Wallace, Bram and Dang, Meihua and Rafailov, Rafael and Zhou, Linqi and Lou, Aaron and Purber, Senthil and Ermon, Stefano and Xiong, Caiming and Joty, Shafiq and Naik, Nikhil},
  journal={arXiv preprint arXiv:2311.12908},
  year={2024}
}

@article{liu2025flow,
  title={Flow-GRPO: Training flow matching models via online RL},
  author={Liu, Jie and Liu, Gongye and Liang, Jiajun and Li, Yangguang and Liu, Jiaheng and Wang, Xintao and Wan, Pengfei and Zhang, Di and Ouyang, Wanli},
  journal={arXiv preprint arXiv:2505.05470},
  year={2025}
}

@article{xue2025dancegrpo,
  title={DanceGRPO: Unleashing GRPO on visual generation},
  author={Xue, Zeyue and Wu, Jie and Gao, Yu and Kong, Fangyuan and Zhu, Lingting and Chen, Mengzhao and Liu, Zhiheng and Liu, Wei and Guo, Qiushan and Huang, Weilin and others},
  journal={arXiv preprint arXiv:2505.07818},
  year={2025}
}

@article{zheng2025diffusionnft,
  title={DiffusionNFT: Online diffusion reinforcement with forward process},
  author={Zheng, Kaiwen and Chen, Huayu and Ye, Haotian and Wang, Haoxiang and Zhang, Qinsheng and Jiang, Kai and Su, Hang and Ermon, Stefano and Zhu, Jun and Liu, Ming-Yu},
  journal={arXiv preprint arXiv:2509.16117},
  year={2025}
}

@article{xu2024imagereward,
  title={ImageReward: Learning and evaluating human preferences for text-to-image generation},
  author={Xu, Jiazheng and Liu, Xiao and Wu, Yuchen and Wang, Yuxuan and Ye, Weiyun and Geng, Shihao and Zhao, Yiren and Li, Jiaming and Li, Cunjian and Sun, Hang and others},
  journal={Advances in Neural Information Processing Systems},
  volume={36},
  year={2023}
}

@article{wu2023human,
  title={Human preference score v2: A solid benchmark for evaluating human preferences of text-to-image synthesis},
  author={Wu, Xiaoshi and Hao, Yiming and Sun, Keqiang and Chen, Yixiong and Zhu, Feng and Zhao, Rui and Li, Hongsheng},
  journal={arXiv preprint arXiv:2306.09341},
  year={2023}
}

@inproceedings{wu2024qbench,
  title={Q-Bench: A benchmark for general-purpose foundation models on low-level vision},
  author={Wu, Haoning and Zhang, Zicheng and Zhang, Erli and Chen, Chaofeng and Liao, Liang and Wang, Annan and Li, Chunyi and Sun, Wenxiu and Yan, Qiong and Zhai, Guangtao and others},
  booktitle={ICLR},
  year={2024}
}

@article{zhang2024qbench,
  title={Q-Bench: A benchmark for multi-modal foundation models on low-level vision from single images to pairs},
  author={Zhang, Zicheng and Wu, Haoning and Zhang, Erli and Zhai, Guangtao and Lin, Weisi},
  journal={IEEE Transactions on Pattern Analysis and Machine Intelligence},
  year={2024}
}

@inproceedings{chen2024mllm,
  title={MLLM-as-a-Judge: Assessing multimodal LLM-as-a-judge with vision-language benchmark},
  author={Chen, Dongping and Chen, Ruoxi and Zhang, Shilin and Wang, Yaochen and Liu, Yinuo and Zhou, Huichi and Zhang, Qihui and Wan, Yao and Zhou, Pan and Sun, Lichao},
  booktitle={Forty-first International Conference on Machine Learning},
  year={2024}
}

@article{wang2025unified,
  title={Unified reward model for multimodal understanding and generation},
  author={Wang, Yibin and Zang, Yuhang and Li, Hao and Jin, Cheng and Wang, Jiaqi},
  journal={arXiv preprint arXiv:2503.05236},
  year={2025}
}

@article{gong2025onereward,
  title={OneReward: Unified mask-guided image generation via multi-task human preference learning},
  author={Gong, Yuan and Wang, Xionghui and Wu, Jie and Wang, Shiyin and Wang, Yitong and Wu, Xinglong},
  journal={arXiv preprint arXiv:2508.21066},
  year={2025}
}

@article{wu2025qwen,
  title={Qwen-Image technical report},
  author={Wu, Chenfei and Li, Jiahao and Zhou, Jingren and Lin, Junyang and Gao, Kaiyuan and Yan, Kun and Yin, Sheng-ming and Bai, Shuai and Xu, Xiao and Chen, Yilei and others},
  journal={arXiv preprint arXiv:2508.02324},
  year={2025}
}

@inproceedings{heusel2017gans,
  title={GANs trained by a two time-scale update rule converge to a local Nash equilibrium},
  author={Heusel, Martin and Ramsauer, Hubert and Unterthiner, Thomas and Nessler, Bernhard and Hochreiter, Sepp},
  booktitle={Advances in Neural Information Processing Systems},
  volume={30},
  year={2017}
}

@inproceedings{zhang2018unreasonable,
  title={The unreasonable effectiveness of deep features as a perceptual metric},
  author={Zhang, Richard and Isola, Phillip and Efros, Alexei A and Shechtman, Eli and Wang, Oliver},
  booktitle={Proceedings of the IEEE Conference on Computer Vision and Pattern Recognition},
  pages={586--595},
  year={2018}
}

@article{wang2004image,
  title={Image quality assessment: From error visibility to structural similarity},
  author={Wang, Zhou and Bovik, Alan C and Sheikh, Hamid R and Simoncelli, Eero P},
  journal={IEEE Transactions on Image Processing},
  volume={13},
  number={4},
  pages={600--612},
  year={2004}
}

@article{liu2022flow,
  title={Flow straight and fast: Learning to generate and transfer data with rectified flow},
  author={Liu, Xingchao and Gong, Chengyue and Liu, Qiang},
  journal={The International Conference on Learning Representations},
  year={2022}
}

@article{gunjal2025rubrics,
  title   = {Rubrics as Rewards: Reinforcement Learning Beyond Verifiable Domains},
  author  = {Gunjal, Anisha and Wang, Anthony and Lau, Elaine and Nath, Vaskar and He, Yunzhong and Liu, Bing and Hendryx, Sean},
  journal = {arXiv preprint arXiv:2507.17746},
  year    = {2025}
}

@techreport{uniworldv2,
  title       = {{UN I W O R L D-V2: Reinforce Image Editing with Diffusion Negative-Aware Finetuning and MLLM Implicit Feedback}},
  author      = {{UniWorld Team}},
  institution = {Peking University, Rabbitpre AI},
  year        = {2025},
  note        = {Technical Report, arXiv:2510.16888}
}

@article{li2025dit,
  title={Dit-vton: Diffusion transformer framework for unified multi-category virtual try-on and virtual try-all with integrated image editing},
  author={Li, Qi and Qiu, Shuwen and Han, Julien and Xu, Xingzi and Seyfioglu, Mehmet Saygin and Koo, Kee Kiat and Bouyarmane, Karim},
  journal={arXiv preprint arXiv:2510.04797},
  year={2025}
}

@article{seyfioglu2024diffuse,
  title={Diffuse to choose: Enriching image conditioned inpainting in latent diffusion models for virtual try-all},
  author={Seyfioglu, Mehmet Saygin and Bouyarmane, Karim and Kumar, Suren and Tavanaei, Amir and Tutar, Ismail B},
  journal={arXiv preprint arXiv:2401.13795},
  year={2024}
}

@article{ghosh2023geneval,
  title={Geneval: An object-focused framework for evaluating text-to-image alignment},
  author={Ghosh, Dhruba and Hajishirzi, Hannaneh and Schmidt, Ludwig},
  journal={Advances in Neural Information Processing Systems},
  volume={36},
  pages={52132--52152},
  year={2023}
}

@inproceedings{hu2023tifa,
  title={Tifa: Accurate and interpretable text-to-image faithfulness evaluation with question answering},
  author={Hu, Yushi and Liu, Benlin and Kasai, Jungo and Wang, Yizhong and Ostendorf, Mari and Krishna, Ranjay and Smith, Noah A},
  booktitle={Proceedings of the IEEE/CVF International Conference on Computer Vision},
  pages={20406--20417},
  year={2023}
}

@misc{shao2026drtulureinforcementlearning,
      title={DR Tulu: Reinforcement Learning with Evolving Rubrics for Deep Research},
      author={Rulin Shao and Akari Asai and Shannon Zejiang Shen and Hamish Ivison and Varsha Kishore and Jingming Zhuo and Xinran Zhao and Molly Park and Samuel G. Finlayson and David Sontag and Tyler Murray and Sewon Min and Pradeep Dasigi and Luca Soldaini and Faeze Brahman and Wen-tau Yih and Tongshuang Wu and Luke Zettlemoyer and Yoon Kim and Hannaneh Hajishirzi and Pang Wei Koh},
      year={2026},
      eprint={2511.19399},
      archivePrefix={arXiv},
      primaryClass={cs.CL},
      url={https://arxiv.org/abs/2511.19399},
      note={Accessed: 2026-06-30},
}

@article{labs2025flux,
  title={FLUX. 1 Kontext: Flow Matching for In-Context Image Generation and Editing in Latent Space},
  author={Labs, Black Forest and Batifol, Stephen and Blattmann, Andreas and Boesel, Frederic and Consul, Saksham and Diagne, Cyril and Dockhorn, Tim and English, Jack and English, Zion and Esser, Patrick and others},
  journal={arXiv preprint arXiv:2506.15742},
  year={2025}
}

@inproceedings{vton-vllm2025,
title={{VTON-VLLM}: Aligning Virtual Try-On Models with Human Preferences},
author={Siqi Wan and Jingwen Chen and Qi Cai and Yingwei Pan and Ting Yao and Tao Mei},
booktitle={NeurIPS},
year={2025}
}

@article{catvton,
  title={Catvton: Concatenation is all you need for virtual try-on with diffusion models},
  author={Chong, Zheng and Dong, Xiao and Li, Haoxiang and Zhang, Shiyue and Zhang, Wenqing and Zhang, Xujie and Zhao, Hanqing and Jiang, Dongmei and Liang, Xiaodan},
  journal={arXiv preprint arXiv:2407.15886},
  year={2024}
}

@article{catv2ton,
  title={Catv2ton: Taming diffusion transformers for vision-based virtual try-on with temporal concatenation},
  author={Chong, Zheng and Zhang, Wenqing and Zhang, Shiyue and Zheng, Jun and Dong, Xiao and Li, Haoxiang and Wu, Yiling and Jiang, Dongmei and Liang, Xiaodan},
  journal={arXiv preprint arXiv:2501.11325},
  year={2025}
}

@inproceedings{catdm,
  title={Cat-dm: Controllable accelerated virtual try-on with diffusion model},
  author={Zeng, Jianhao and Song, Dan and Nie, Weizhi and Tian, Hongshuo and Wang, Tongtong and Liu, An-An},
  booktitle={Proceedings of the IEEE/CVF conference on computer vision and pattern recognition},
  pages={8372--8382},
  year={2024}
}

@inproceedings{idmvton,
  title={Improving diffusion models for authentic virtual try-on in the wild},
  author={Choi, Yisol and Kwak, Sangkyung and Lee, Kyungmin and Choi, Hyungwon and Shin, Jinwoo},
  booktitle={European Conference on Computer Vision},
  pages={206--235},
  year={2024},
  organization={Springer}
}

@misc{ootdiffusion,
      title={OOTDiffusion: Outfitting Fusion based Latent Diffusion for Controllable Virtual Try-on}, 
      author={Yuhao Xu and Tao Gu and Weifeng Chen and Chengcai Chen},
      year={2024},
      eprint={2403.01779},
      archivePrefix={arXiv},
      primaryClass={cs.CV},
      url={https://arxiv.org/abs/2403.01779},
      note={Accessed: 2026-06-30},
}

@inproceedings{omnivton,
  title={Omnivton: Training-free universal virtual try-on},
  author={Yang, Zhaotong and Li, Yuhui and He, Shengfeng and Li, Xinzhe and Xu, Yangyang and Dong, Junyu and Du, Yong},
  booktitle={Proceedings of the IEEE/CVF International Conference on Computer Vision},
  pages={16702--16711},
  year={2025}
}

@article{rubicon2025,
  title={Reinforcement learning with rubric anchors},
  author={Huang, Zenan and Zhuang, Yihong and Lu, Guoshan and Qin, Zeyu and Xu, Haokai and Zhao, Tianyu and Peng, Ru and Hu, Jiaqi and Shen, Zhanming and Hu, Xiaomeng and others},
  journal={arXiv preprint arXiv:2508.12790},
  year={2025}
}

@article{zhang2025treegrpo,
  title={TreeGRPO: Tree-Advantage GRPO for Online RL Post-Training of Diffusion Models},
  author={Ding, Zheng and Ye, Weirui},
  journal={arXiv preprint arXiv:2512.08153},
  year={2025}
}

@article{wang2025grpoguard,
  title={Grpo-guard: Mitigating implicit over-optimization in flow matching via regulated clipping},
  author={Wang, Jing and Liang, Jiajun and Liu, Jie and Liu, Henglin and Liu, Gongye and Zheng, Jun and Pang, Wanyuan and Ma, Ao and Xie, Zhenyu and Wang, Xintao and Wang, Meng and Wan, Pengfei and Liang, Xiaodan},
  journal={arXiv preprint arXiv:2510.22319},
  year={2025}
}

@article{cocoedit2025,
  title={CoCoEdit: Content-Consistent Image Editing via Region Regularized Reinforcement Learning},
  author={Wu, Yuhui and Xie, Chenxi and Li, Ruibin and Chen, Liyi and Yi, Qiaosi and Zhang, Lei},
  journal={arXiv preprint arXiv:2602.14068},
  year={2026}
}

@article{li2025mixgrpo,
  title={Mixgrpo: Unlocking flow-based grpo efficiency with mixed ode-sde},
  author={Li, Junzhe and Cui, Yutao and Huang, Tao and Ma, Yinping and Fan, Chun and Yang, Miles and Zhong, Zhao},
  journal={arXiv preprint arXiv:2507.21802},
  year={2025}
}

@article{he2025superflow,
  title={SuperFlow: Training Flow Matching Models with RL on the Fly},
  author={Chen, Kaijie and Xu, Zhiyang and Shen, Ying and Lin, Zihao and Yao, Yuguang and Huang, Lifu},
  journal={arXiv preprint arXiv:2512.17951},
  year={2025}
}

@article{jacob2025qa,
  title={Qa-lign: Aligning llms through constitutionally decomposed qa},
  author={Jacob Dineen, Aswin RRV and Liu, Qin and Xu, Zhikun and Ye, Xiao and Shen, Ming and Li, Zhaonan and Lu, Shijie and Baral, Chitta and Chen, Muhao and Zhou, Ben},
  journal={arXiv preprint arXiv:2506.08123},
  year={2025}
}
